\documentclass[runningheads]{llncs}

\usepackage[mobile]{eccv}

\usepackage{graphicx}
\usepackage{amsmath}
\usepackage{amssymb}
\usepackage{booktabs}
\usepackage{tabularx}
\usepackage{multirow}
\newcolumntype{Y}{>{\centering\arraybackslash}X}
\usepackage{xcolor}

\usepackage{pifont}
\usepackage[bottom]{footmisc}
\usepackage{algorithm}
\usepackage{comment}
\usepackage{algpseudocode}
\usepackage{float}

\usepackage[accsupp]{axessibility}  

\usepackage{amsthm}

\newtheorem{theo}{Theorem}

\newcommand{\R}{\mathbb{R}}
\let \bs=\mathbf
\let \set=\mathcal

\def \diag {\mathrm{diag}}

\def \path {\mathit{path}}

\def \exp {\textup{exp}}

\newtheorem{lem}{\textbf{Lemma}}

\let \set = \mathcal
\let \bs = \boldsymbol

\newcommand\norm[1]{\lVert#1\rVert}

\usepackage[pagebackref,breaklinks,colorlinks,citecolor=eccvblue]{hyperref}

\usepackage[capitalize]{cleveref}
\crefname{section}{Sec.}{Secs.}
\Crefname{section}{Section}{Sections}
\Crefname{table}{Table}{Tables}
\crefname{table}{Tab.}{Tabs.}

\usepackage{orcidlink}

\begin{document}

\title{Freditor: High-Fidelity and Transferable NeRF Editing by Frequency Decomposition} 

\titlerunning{Freditor}

\newcommand\blfootnote[1]{%
  \begingroup
  \renewcommand\thefootnote{}\footnote{#1}%
  \addtocounter{footnote}{-1}%
  \endgroup
}

\author{Yisheng He\inst{1}$^\ast$ \and Weihao Yuan\inst{1}$^\ast$ \and Siyu Zhu\inst{2} \and Zilong Dong\inst{1} \and Liefeng Bo\inst{1} \and Qixing Huang\inst{3}}

\authorrunning{He et al.}

\institute{Alibaba Group \and Fudan University \and The University of Texas at Austin}

\maketitle

\blfootnote{$^\ast$Equal contributions.}%

\begin{figure}[ht] 
    \centering
    \includegraphics[width=1.\textwidth]{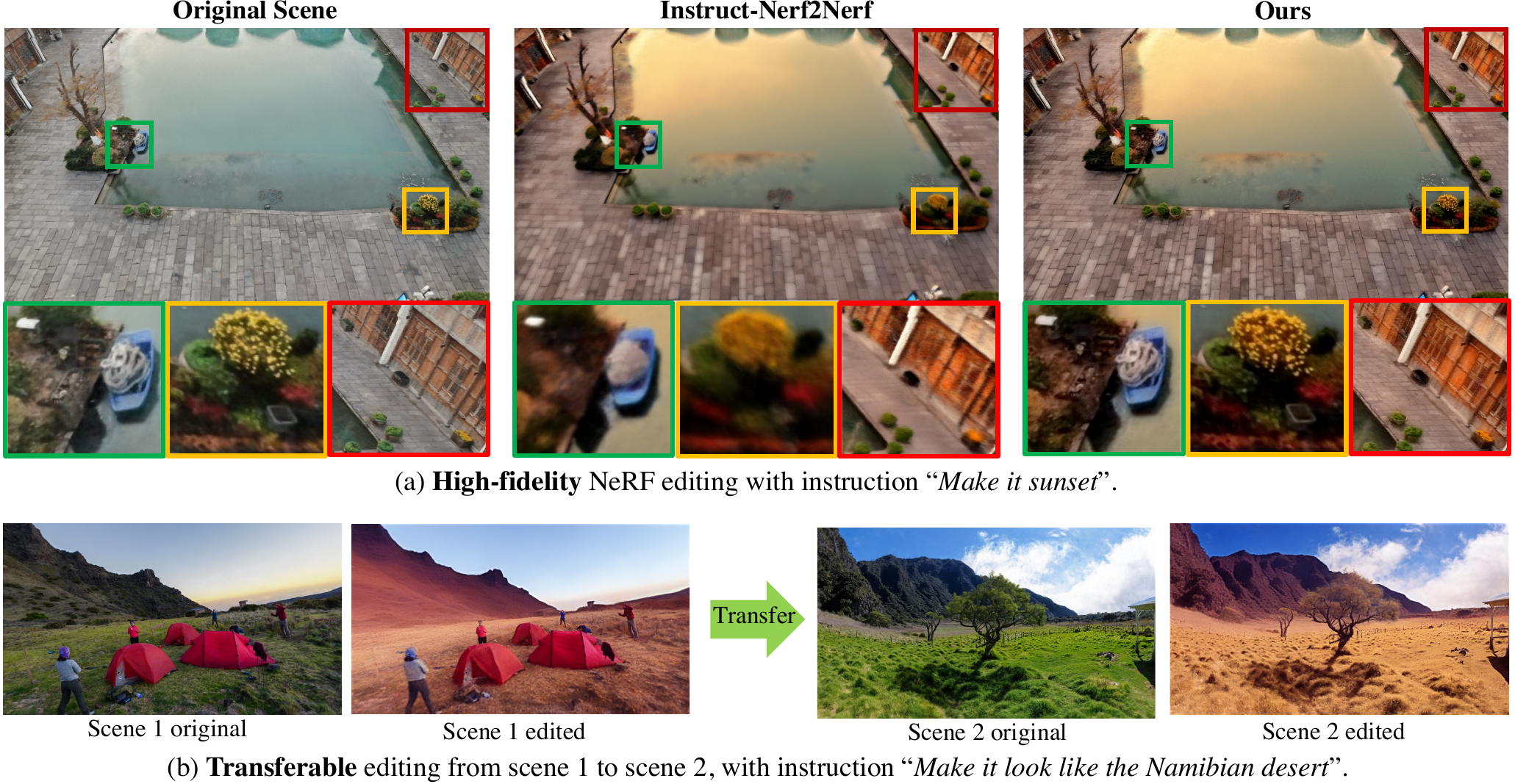} 
    \vspace{-4mm}
    \caption{High-fidelity and transferable 3D scenes editing with text instructions. (a) High-fidelity scene editing compared to Instruct-NeRF2NeRF is displayed. Three color patches are zoomed in to highlight the details. (b) The editing trained in one scenario could be directly transferred to different novel scenarios without the need for retraining.}
    \label{fig:teaser}
    \vspace{-7mm}
\end{figure}
\begin{abstract}

This paper enables high-fidelity, transferable NeRF editing by frequency decomposition. 
Recent NeRF editing pipelines lift 2D stylization results to 3D scenes while suffering from blurry results, and fail to capture detailed structures caused by the inconsistency between 2D editings. 
Our critical insight is that low-frequency components of images are more multiview-consistent after editing compared with their high-frequency parts. Moreover, the appearance style is mainly exhibited on the low-frequency components, and the content details especially reside in high-frequency parts. 
This motivates us to perform editing on low-frequency components, which results in high-fidelity edited scenes.
In addition, the editing is performed in the low-frequency feature space, enabling stable intensity control and novel scene transfer.
Comprehensive experiments conducted on photorealistic datasets demonstrate the superior performance of high-fidelity and transferable NeRF editing. The project page is at \url{https://aigc3d.github.io/freditor}.

\end{abstract}
\section{Introduction}

The manipulation and enhancement of three-dimensional environments through 3D scene editing is a fundamental task in the field of computer vision, with numerous practical applications in areas such as augmented reality, autonomous driving, and virtual simulation, since reconstructing a 3D scene is laborious and time-consuming, such as recovering a scene of different seasons.
This paper, therefore, studies the problem of high-fidelity and transferable photo-realistic editing of 3D scenes with text instructions.
With the emergence of the neural radiance field (NeRF), the photorealistic reconstruction of a 3D scene from 2D images has been significantly simplified.
Thus, we employ NeRF as our 3D modeling in this work.
Given a NeRF representing an underlying 3D scene, our objective is to edit this scene according to text instructions such that the edited NeRF can render images with instructed styles that differ from the captured images.
This language-based interface enables easy editing of 3D content.

Previous methods could directly perform 2D editing on the rendered image sequence~\cite{img2_Gatys_Ecker_Bethge_2016,Huang_Belongie_2017,Johnson_Alahi_Fei-Fei_2016} from NeRF. However, existing studies~\cite{Huang_Tseng_Saini_Singh_Yang_2021,refart3_Huang_He_Yuan_Lai_Gao,Mu_Wang_Wu_Li_2022,refart4_Nguyen-Phuoc_Liu_Xiao} have illustrated that simply combining 3D novel view synthesis and 2D style transfer often results in multi-view inconsistency or subpar stylization quality. To overcome these limitations, it is necessary to jointly optimize novel view synthesis and style transfer for 3D scenes.
Recent research like Instruct-NeRF2NeRF~\cite{haque2023instructnerf2nerf} shows the possibility of retraining the NeRF with edited images.
However, such a pipeline has several limitations. 
First, the edited scene is blurry because of the view inconsistency caused by perturbations of the 2D stable diffusion models. Second, it requires training for each scene and each style, which is time-consuming and hinders its ability to scale up. These are two common problems in modern 3D content stylization frameworks~\cite{haque2023instructnerf2nerf,light1_Verbin_Hedman_Mildenhall_Zicklen_2022,climatenerf}.

\begin{figure}[htbp]
\centering
\begin{minipage}[ht]{0.49\linewidth}
\centering
\includegraphics[width=1\linewidth]{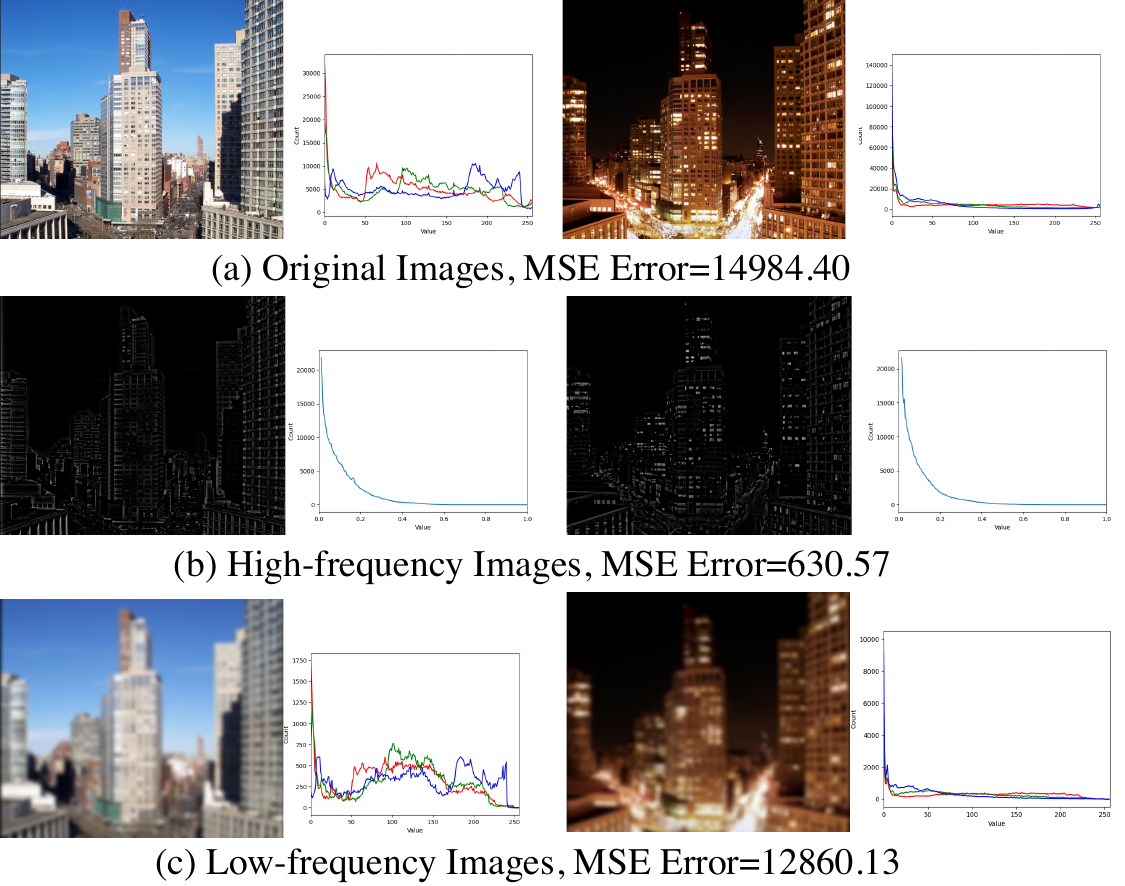}
\caption{Visualization of the difference between two styles in different frequency images. 
Two styles of the same scene are displayed. 
The MSE error between the two styles is computed, and the RGB distribution curves are placed close to each image. We can see the discrepancy between different styles in high-frequency is minor while that in low-frequency is significant.
}
\label{fig:low_high_diff}
\end{minipage}
\hspace{1mm}
\begin{minipage}[ht]{0.48\linewidth}
\centering
\includegraphics[width=0.9\linewidth]{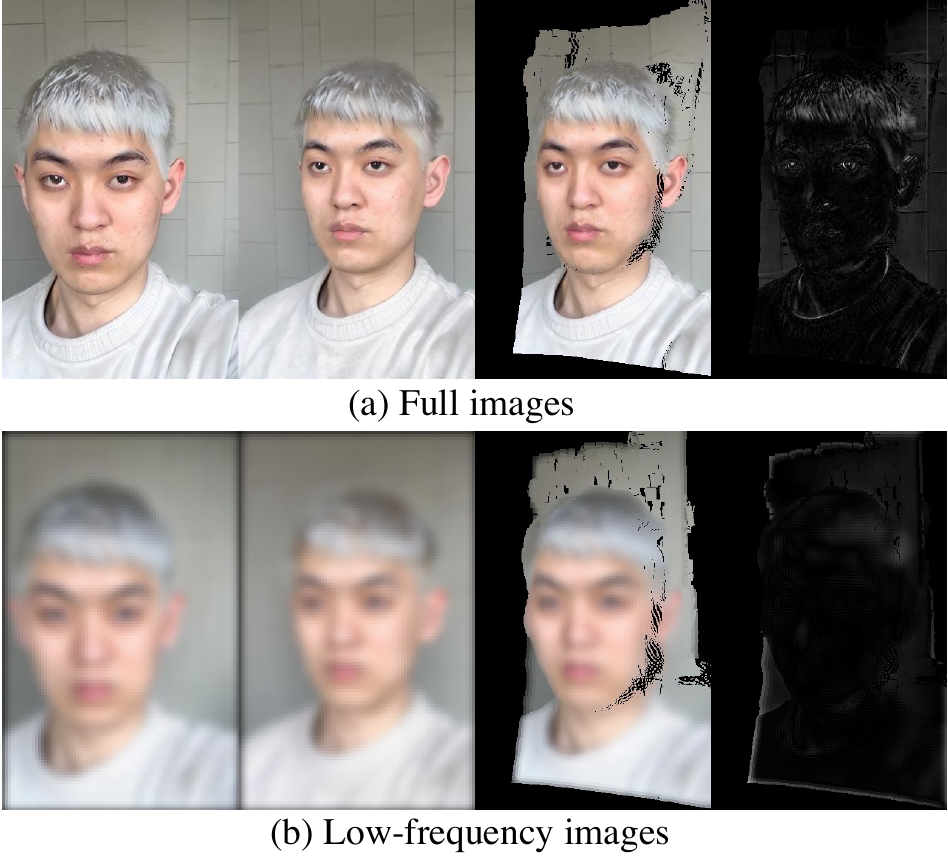}
\caption{
Visualization of inconsistency regions caused by 2D editing. Images from left to right represent the edited view 1, view 2, view 1 wrapped to view 2 by optical-flow algorithm, and the inconsistency map between the two views. We can see that the inconsistency in edited high-frequency details is larger than the low-frequency ones. 
}
\label{fig:low_high_cmp}
\end{minipage}
\end{figure}

To address these challenges, in this paper, we endeavor to facilitate high-fidelity and transferable NeRF stylization by decomposing the appearance in the frequency domain.
Our critical insight is that the appearance style manifests itself predominantly in the low-frequency components, and the content details reside especially in the high-frequency parts~\cite{liang2021highresollaplacian} including both the geometry details and the texture details, as shown in Figure~\ref{fig:low_high_diff}. On the other hand, the low-frequency components of images, which predominantly define the appearance style, exhibit enhanced multi-view consistency after editing compared to their high-frequency counterparts.
This motivates us to do style editing on low-frequency components and transfer the high-frequency information back to the edited scene, which yields a high-resolution edited scene. 
This pipeline additionally makes it easier for a style network to transfer across different scenes and styles, since the low-frequency components contain fewer scene-dependent details to preserve.

Starting from the frequency decomposition, we build our feature-space NeRF editing framework, as presented in Figure~\ref{fig:framework}.
The high-frequency branch is taken care of by a standard NeRF in color-space rendering, while the low-frequency branch is taken care of in feature-space rendering, which enables easy low-frequency extraction and style editing.
The raw feature or the edited feature can then be decoded by the same decoder, producing the raw low-frequency and stylized low-frequency components, respectively.
Benefiting from the decomposition, we could relieve the inconsistency between edited views and recover the high-frequency details after the editing.
Benefiting from the feature manipulation, we could control the editing level by interpolating in the feature space.
This controllable editing is more stable than previous methods, due to the variability of high-frequency signals addressed in our framework.

Moreover, the trained stylization could be directly transferred to a new scenario without any retraining.
The stylization module could be even trained in 2D images and applied to 3D scenes, further alleviating the 3D editing workload.

In the experiments, we demonstrate our method produces consistent and high-fidelity appearances of edited 3D scenes compared to previous methods.
Also, our experiments illustrate the capability of intensity control and style transfer which are absent in the compared methods.

The main contributions of this work are summarized in the following.

$\bullet$ We decompose the low-frequency and high-frequency components in the NeRF rendering, such that the editing could be performed in the low-frequency space to better keep the multi-view consistency, and the appearance details could be retained in the high-frequency space.

$\bullet$ The editing is performed by a stylizer on the feature map, which is decoded by a shared decoder to the rendered color. Therefore, the intensity of the stylization is controllable by interpolating between the edited feature and the original feature in low-frequency space, which is particularly stable due to in low-frequency space.

$\bullet$ The trained stylization is transferable to a new scenario without the need for retraining. Furthermore, we could train a stylization on a collection of 2D images and apply it to a 3D NeRF scenario.

\section{Related Work}

\subsection{Physical-based NeRF editing} 
Physical-based approaches~\cite{light1_Verbin_Hedman_Mildenhall_Zicklen_2022,light2_Boss_Jampani_Braun_Liu,light3_Mildenhall_Barron_2021,light4_Munkberg_Hasse,light5-Brualla_Srinivasan_Barron_2022,PIE-NeRF,PhysGaussian} impose physics-based biases, e.g, lighting and material~\cite{light1_Verbin_Hedman_Mildenhall_Zicklen_2022,light2_Boss_Jampani_Braun_Liu,light3_Mildenhall_Barron_2021,light4_Munkberg_Hasse,light5-Brualla_Srinivasan_Barron_2022} or use bounding boxes~\cite{box1,box2} for specific edits like object deformation~\cite{box_deform}. ClimateNeRF~\cite{climatenerf} combines the physical engine to simulate weather changes that are supported by the engine, e.g. snow and flood. 
Although the target weathers are achieved by the simulation, a physical engine is needed for each style, which is too heavy to scale up to all kinds of scenarios.
In contrast, built on~\cite{haque2023instructnerf2nerf}, our work enables more flexible edits in NeRFs, e.g., any global style changing or local editing of the specific object region without relying on physical properties or simulations. 

\subsection{Artistic NeRF Editing} 
Methods working on artistic NeRF editing~\cite{refart1_Chiang_2021,Huang_Tseng_Saini_Singh_Yang_2021,refart3_Huang_He_Yuan_Lai_Gao,refart4_Nguyen-Phuoc_Liu_Xiao,refart5_Zhang_Kolkin_Bi,refart6_Wu_Tan_Xu_2022,paintnesf} draw inspiration from image stylization~\cite{imgart1_Hertzmann_1998,img2_Gatys_Ecker_Bethge_2016} and focus on the 3D artistic stylization of NeRFs using a 2D reference artistic image. 
Recently, StyleRF~\cite{liu2023stylerf} introduces feature-based zero-shot artistic style editing, but it mainly addresses global scene appearance changes and produces nonphotorealistic results. In contrast, our work leverages frequency decomposition techniques to enable high-fidelity and transferable photorealistic style editing.

\subsection{Text-Driven NeRF Editing}
Text-driven techniques have been integrated into NeRF stylization to facilitate convenient and user-friendly editing~\cite{clip_Radford_Kim,haque2023instructnerf2nerf,Lseg_Li_Weinberger_Belongie_Koltun_Ranftl,SKED,FaceCLIPNeRF,Blending-NeRF}. An approach involves leveraging CLIP-based latent representations~\cite{clip_Radford_Kim}. EditNeRF~\cite{EditNeRF_Liu_Zhang_Zhang_Zhang_Zhu_Russell_2021} focuses on manipulating latent codes from synthetic datasets, while ClipNeRF~\cite{ClipNeRF_Wang_Chai_He_Chen_Liao_2022} and NeRF-Art~\cite{NeRFArt_Wang_Jiang_Chai_He_Chen_Liao_2022} incorporate text prompts. However, these CLIP-based methods have limitations when it comes to localized edits. Other approaches, such as~\cite{clipdino_Kobayashi_Matsumoto_Sitzmann,clipdino_Tschernezki_Laina_Larlus_Vedaldi}, distill DINO~\cite{DINO_Caron_Touvron_Misra_Jegou_Mairal_Bojanowski_Joulin_2021} or Lseg~\cite{Lseg_Li_Weinberger_Belongie_Koltun_Ranftl} features to enable localized editing techniques such as 3D spatial transformations~\cite{ClipNeRF_Wang_Chai_He_Chen_Liao_2022} and object removal~\cite{clipdino_Tschernezki_Laina_Larlus_Vedaldi}. Instruct-NeRF2NeRF~\cite{haque2023instructnerf2nerf} and some follow-ups~\cite{GaussianEditor2,GaussianEditor,PDS,InstructPix2NeRF} elevate instruction-based 2D image-conditioned diffusion models to 3D and allow a purely language-based interface for local and global editing. However, they require training for each scene and style, and inconsistent 2D editing can result in blurry artifacts in the edited results. To address these limitations, we introduce the frequency decomposition technique on the feature space, enabling transferable and high-fidelity NeRF editing.

\section{Approach}

\subsection{Overview}
\label{Subsec:App:Overview}

Our goal is to achieve a series of sequential tasks, namely per-scene stylization, controllable stylization, and transferable stylization.

\noindent\textbf{Per-scene editing:} 
Given a set of posed images $\set{I} = \{I_1, ..., I_{N}\}$ of an underlying 3D scene, a reconstructed 3D scene representation from $\set{I}$, and an editing instruction $\mathcal{S}$, the goal is to retrain the 3D scene with the text instruction and generate edited novel views $\{I'_1, ..., I'_K\}$ of the edited 3D scene from any camera poses $\{C'_1, ..., C'_K\}$. 

\noindent\textbf{Intensity control:}
After the editing of a 3D scene, the novel views of different editing levels $\{{I'}^l_k\}$ could be directly rendered given the indicator $l \in [0,1]$, where $l=0$ indicates no-editing while $l=1$ indicates full-editing.

\noindent\textbf{Stylization transfer:}
After training a style on a set of images, this style could be directly applied to a new 3D scene without retraining the stylization module on that scene.

\begin{figure*}[htb]
\centering
\includegraphics[width=1.\linewidth]{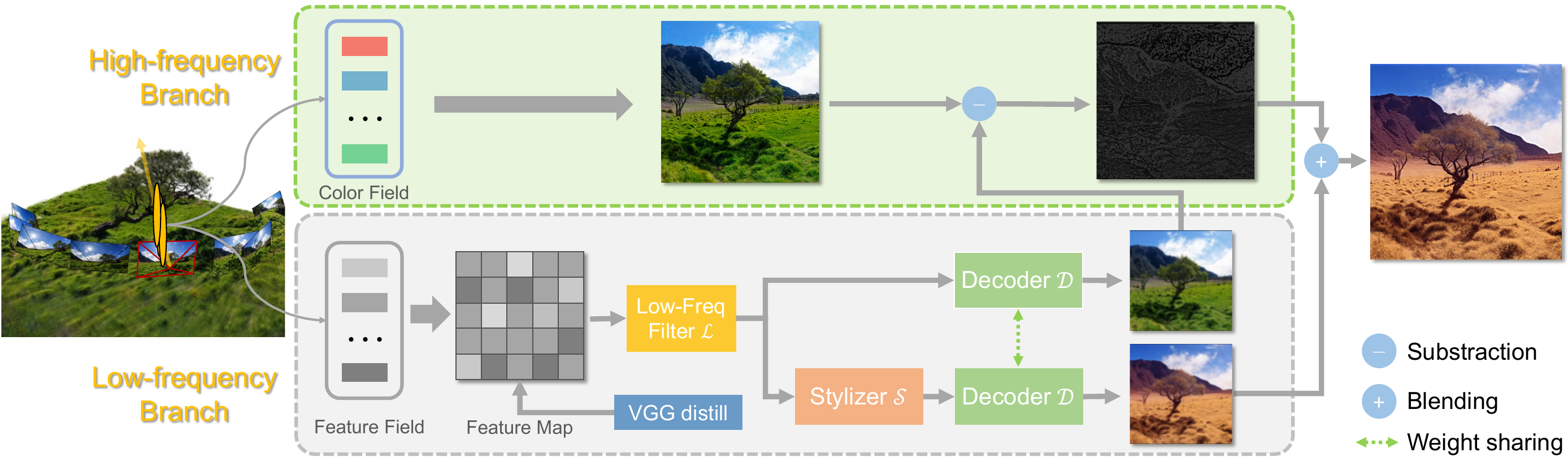}
\caption{\textbf{Overall framework}. Our pipeline comprises two primary branches: the high-frequency branch, reconstructed from multiview images, which ensures view-consistent scene details, and the low-frequency branch, responsible for filtering low-frequency components from the full scene feature fields, performing style transfer, and decoding the original and edited low-frequency images. Finally, the high-frequency details are reintegrated into the edited low-frequency image, resulting in a high-fidelity edited scene.}
\label{fig:framework}
\end{figure*}

We propose a new frequency-decomposed stylization network to achieve the aforementioned goals, as depicted in Figure~\ref{fig:framework}. Our framework comprises two main branches: a high-frequency branch dedicated to preserving content details and a low-frequency branch designed for transferable and controllable stylization. In the high-frequency branch, we employ a vanilla NeRF to reconstruct the highly detailed original scene. To achieve controllable and transferable stylization, the second branch performs low-frequency filtering and editing in the feature space. Alongside RGB values, our neural radiance field also generates a feature map of the original scene. The low-frequency branch, built on this feature map, consists of three modules. 
First, a low-pass filter is used to extract the low-frequency feature from the full scene. 
Next, a stylization network edits the scene in the feature space according to the desired style. 
Then, a shared decoder reconstructs the original and edited low-frequency images from the original and styled low-frequency feature maps, respectively. 
Ultimately, the high-frequency component of the original scene is obtained by subtracting the reconstructed low-frequency image from the original image. By blending the high-frequency details from the original scene with the edited low-frequency component, we obtain a high-fidelity edited image.


\subsection{Consistent Editing by Frequency Decomposition}
\label{Subsec:3.3}
In the pipeline of lifting 2D styled results to a 3D scene, such as Instruct-NeRF2NeRF~\cite{haque2023instructnerf2nerf}, the primary cause of defects and blurring in edited 3D scenes is the inconsistency among different 2D edited views. To address this challenge, we make a key observation that editing in the low-frequency space provides greater ease in preserving multiview consistency, as shown in Figure~\ref{fig:low_high_cmp}. On the basis of this insight, we developed the frequency-decomposed stylization network. In the following section, we present evidence supporting this observation.

To begin, we implement the decomposition of an image into its low- and high-frequency components. This decomposition process is achieved through the utilization of a low-pass filter commonly used in image processing, such as the Gaussian filter. Specifically, the filter extracts the low-frequency components from the original images. By subtracting these low-frequency components from the original image, we are left with the high-frequency details.

Denote $\set{L}: \R^{n\times n}\rightarrow \R^{n\times n}$ as the low-pass filter that performs image smoothing. Let $\set{A}:\R^{n\times n} \rightarrow \R^{n\times n}$ be the stylization operation. Without losing generality, we assume
$$
\set{A}\circ \set{L} \approx \set{L}\circ \set{A},
$$
which means we get similar results when stylizing the smoothed image and smoothing the stylized image. This assumption is applicable to most stylization and smoothing operations.


The following theorem, which is stated informally, shows that editing in the low-frequency component leads to more consistent editing. The formal statement is deferred to the supp. material.

\begin{theo}
(\textbf{Informal}) Consider an optical flow-based consistency score $c(I_1, I_2)$ between images $I_1$ and $I_2$, we have
\begin{equation}
c\big(\set{A}(\set{L}(I_1)), \set{A}(\set{L}(I_2))\big) < c\big(\set{A}(I_1), \set{A}(I_2)\big).
\label{eqn:consistency}
\end{equation}
\end{theo}

\noindent\textsl{Sketch Proof:}
We first show that for a smoothing operator $\set{L}$,
$$
c(\set{L}(I_1),\set{L}(I_2)) < c(I_1, I_2).
$$
As $\set{A} \circ \set{L} \approx \set{L} \circ \set{A} $, we have 
\begin{equation*}
\begin{split}
c\big(\set{A}(\set{L}(I_1)), \set{A}(\set{L}(I_2))\big)
= &\ \ c((\set{A}\circ \set{L})(I_1),(\set{A}\circ \set{L})(I_2)) \\
\approx  & \ \ c((\set{L}\circ \set{A})(I_1),(\set{L}\circ \set{A})(I_2))\\
< & \ \ c (\set{A}(I_1), \set{A}(I_2)).
\end{split}.
\end{equation*}
The details are deferred to the supp. material. 



Based on this theoretical insight, we proceed to introduce our consistent 3D stylization networks based on frequency decomposition.

\subsection{Frequency Decomposed NeRF Editing in Feature Space}
\label{Subsec:3.4}

\paragraph{Preliminaries: NeRF.} NeRF~\cite{DBLP:conf/eccv/MildenhallSTBRN20} is a computational representation for complex 3D scenes, designed to capture complete volumetric scene information by providing a dense understanding of the photometric properties of the scene.
Developed as a fully connected neural network, NeRF synthesizes new 2D images through intricate volumetric rendering of scenes based on learned 3D geometry and appearance information.
Given the viewing direction $\mathbf{d}$, for any 3D point $(x, y, z)$ in the 3D scene, the MLPs map the point to its geometry density $\sigma$ and appearance color $\mathbf{c}$.
Then for any pixel of the image to render, the points on the ray $\mathbf{r}$ corresponding to this pixel are accumulated to produce the pixel color $C(\mathbf{r})$ by ray marching $\mathbf{r}(t)=\mathbf{o}+t\mathbf{d}$, as
\begin{align}
\begin{aligned}
    C(\mathbf{r}) &= \int_{t_n}^{t_f} T(t)\sigma(\mathbf{r}(t))\mathbf{c}(\mathbf{r}(t),\mathbf{d})dt   \\
    T(t) &= \exp { \Big(-\int_{t_n}^{t_f} \sigma(\mathbf{r}(s))ds \Big)}
\end{aligned},
\end{align}
where $t_n, t_f$ are the near and far bounds, respectively.

\paragraph{Per-scene High-fidelity Editing.}
\label{Subsec:3.4}
A vanilla trial to apply frequency-decomposed stylization on NeRF is to build a NeRF that reconstructs the low- and high-frequency components of the scene using two separate MLP headers. Specifically, one header renders high-frequency pixel colors, and the other renders low-frequency ones. Subsequently, we can perform a consistent multiview style transfer on the low-frequency header and then blend the high-frequency details back. Under this formulation, one practical system built on Instruct-NeRF2NeRF is to regard its MLP header as the blurry low-frequency branch and plug in another branch to learn the high-frequency details of the original scene. The high-frequency details can then be blended back into the edited low-frequency scene to enhance the fidelity. In this way, we can obtain high-fidelity edited scenes.

\paragraph{Feature-space Rendering.} 
\label{Subsec:3.5}


Though getting high-fidelity edited results, the above per-scene editing pipeline still has several limitations. First, the MLPs are learned per scene in NeRF, limiting the ability to transfer the learned style to novel scenes. Second, the intensity of edited results is not controllable once the network is trained. To tackle these limitations, we utilize the feature field, which renders a feature map of the scene. We then perform frequency decomposition and style transfer on this feature map. In this way, the intensity control can be accomplished by feature interpolation and the transferable editing can be achieved by performing the stylization in the same feature domain. 

To obtain the feature field in NeRF, alongside the color of each point, we add another MLP header to output each point's feature $\mathbf{f}$, which is then accumulated to a feature $\mathbf{F}$ that corresponds to the query pixel by ray-marching:
\begin{align}
\begin{aligned}
    F(\mathbf{r}) &= \int_{t_n}^{t_f} T(t)\sigma(\mathbf{r}(t))\mathbf{f}(\mathbf{r}(t),\mathbf{d})dt.
\end{aligned}
\end{align}

\paragraph{Feature-space Decomposation.}
To enable transferable style editing, we enforce the rendered feature space to be in the VGG~\cite{DBLP:journals/corr/SimonyanZ14a} feature field. This ensures that the following modules learned to process in this domain can be shared and generalized across different scenes.
Given the rendered feature map of the scene, we can perform low-frequency filtering, stylization, and low-frequency image reconstruction on it. To begin, we need to extract a low-frequency feature from it and decode the original low-frequency image. Specifically, the low-frequency filter module $\mathcal{L}$ consists of several learnable low-pass filters that extract low-frequency information from the original feature map. 
In practice, we use convolution layers followed by downsampling to implement the low-pass filters. The low-frequency filtered feature is then fed into a convolution decoder $\mathcal{D}$. The decoder reconstructs the original low-frequency image $C_{low}$ from the filtered low-frequency feature map, as
\begin{equation}
    C_{low} = \mathcal{D} (\mathcal{L}(F) ).
\end{equation}

\paragraph{Low-frequency Editing.}
To do style editing on the low-frequency feature map, we install a style module $\mathcal{S}$ between the low-frequency filter and the decoder. This style module is composed of several convolution layers with nonlinear activation. It transfers the low-frequency feature map to a feature map of the target style, and the decoder reconstructs the styled low-frequency image $C_{styledLow}$ as:
\begin{equation}
    C_{styledLow} = \mathcal{D}(\mathcal{S}(\mathcal{L}(F))).
\end{equation}

Given the rendered original image $C$, the original low-frequency component $C_{low}$, and edited low-frequency component $C_{styledLow}$, we blend high-frequency details from the original image to the edited image and obtain a high-resolution edited image $C_{styled}$ as:
\begin{equation}
    C_{styled} = \mathcal{U}(C_{styledLow}) + \alpha (C - \mathcal{U}(C_{low})),
\end{equation}
where $\mathcal{U}$ is a simple bilinear upsampling operation that interpolates the low-frequency images to the size of the original image, and $\alpha \in [0.0, 1.0]$ can adjust the level of high-frequency details blended back to the edited scene.

\subsection{Style Transfer and Intensity Control}

\paragraph{Transfer.} 
We decompose frequencies and edit styles in the distilled VGG feature field, which enables us to transfer the learned low-frequency filter $\mathcal{L}$, stylizer $\mathcal{S}$, and decoder $\mathcal{D}$ from one scene to another. Specifically, For a new scene, we only need to train the high-frequency branch and the feature field. The existing $\mathcal{L}$, $\mathcal{S}$, and $\mathcal{D}$ can then be directly integrated for stylization. This saves time and computational resources as we do not have to retrain these components for each scene separately.

\paragraph{Intensity Control.}
In many real-world applications, it is essential to have the flexibility to control the level of editing during network inference. For instance, scenarios where a gradual transition from day to night is desired. While previous approaches such as Instruct-NeRF2NeRF lack the ability to control the level of editing after training, our framework enables easy control by interpolating between the original and edited low frequency features using an indicator $l \in [0, 1]$. This interpolation process can be expressed as follows:
\begin{equation}
    C_{l StyledLow} = \mathcal{D}(l \mathcal{S}(\mathcal{L}(F)) + (1-l)\mathcal{L}(F)).
\end{equation}

\subsection{Regularizations and Training Pipeline}

The high-frequency branch is trained with captured images by the mean square error (MSE) loss, following common practice for training NeRF models. The low-frequency branch is trained in two stages by first training the frequency decomposition network and then the stylization module.

\paragraph{Stage 1: Training the frequency decomposition network}
In the first stage, we train the feature map renderer, the low-frequency filter $\mathcal{L}$ and the decoder $\mathcal{D}$ with low-frequency ground-truth RGB images and VGG feature maps.
To generate ground-truth low-frequency RGB images, we utilize the commonly used Gaussian filter to extract low-frequency components from the original RGB images. 
To generate ground-truth VGG features, we extract features from the ReLU3-1 layer of a pre-trained VGG and bilinearly upsample that to the full size.

Then the training objective is defined as follows.
\begin{equation}
    l = \norm{F-\hat{F}}^{2}_{2} + \norm{C_{low}-\hat{C_{low}}}^{2}_{2} + \text{lpips}(C_{low}, \hat{C_{low}}),
\end{equation}
where $F$ and $\hat{F}$ denote the rendered and ground-truth VGG features respectively, 
$C_{low}$ and $\hat{C_{low}}$ denote the predicted and ground truth low-frequency image respectively,
and $\text{lpips}()$ denotes the perception loss function following~\cite{Johnson_Alahi_Fei-Fei_2016}.

\paragraph{Stage 2: Training the styling network}
In the second stage, we train the style module $S$. To enable instruction-based photorealistic 3D editing, we transform the 2D editing results from Instruct-Pix2Pix~\cite{brooks2022instructpix2pix} into 3D using an iterative dataset update technique following Instruct-NeRF2NeRF~\cite{haque2023instructnerf2nerf}. Specifically, during training, the images rendered from NeRF will be edited by the diffusion model and subsequently used to supervise the NeRF editing module. One difference is that we only use the low-frequency components of the edited images to train our low-frequency feature stylization module $\mathcal{S}$, which ensures better multiview consistency. Moreover, all other modules, e.g., $\mathcal{L}$ and $\mathcal{D}$, are frozen during this stage. Such a training scheme enables smooth progress in transferring the original low-frequency feature to the styled one. It also enables easy control of the stylization level during inference. The training objective is the MSE loss and perception loss between the predicted and ground-truth low-frequency styled images, as
\begin{equation}
\begin{split}
    l_{style} = &\norm{C_{styledLow} - \hat{C_{styledLow}}}^{2}_{2} \\
                   & + \text{lpips}(C_{styledLow}, \hat{C_{styledLow}}),
\end{split}
\end{equation}
where $C_{styledLow}$ and $\hat{C_{styledLow}}$ denote the predicted and ground truth styled low-frequency image respectively.

\begin{figure*}[t!]
\centering
\includegraphics[width=1\linewidth]{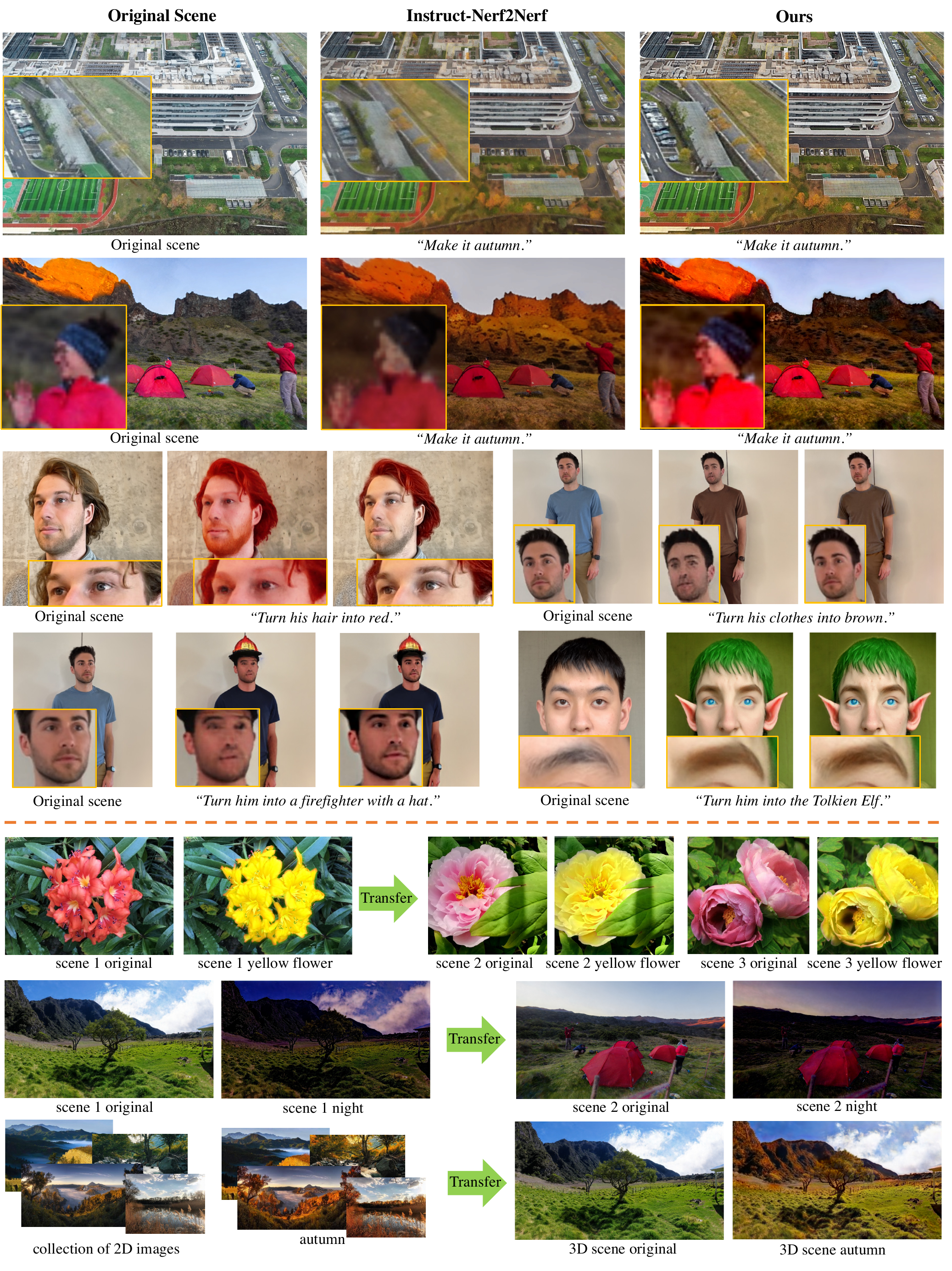}
\vspace{-3mm}
\caption{\textbf{Qualitative Results:} 
Our method can perform high-fidelity edits on real scenes, including environmental changes like adjusting the season, weather, or time, and customized changes of an object or a person in a scene. 
Moreover, the editing trained on scene 1 could be directly transferred to a novel scene 2, and the editing trained on a collection of 2D images could be transferred to 3D scenes.
}
\vspace{-3mm}
\label{fig:vis_res_all}
\end{figure*}
\section{Experiments}

\subsection{Datasets}

We perform the experiments on a total of 10 real scenes taken from previously published datasets~\cite{haque2023instructnerf2nerf,barron2022mipnerf360} and two large-scale scenes captured by a quadcopter drone. In particular, 6 scenes from Instruct-NeRF2NeRF~\cite{haque2023instructnerf2nerf}, 2 scenes from Mip-NeRF 360~\cite{barron2022mipnerf360} are evaluated.
For a fair comparison to Instruct-NeRF2NeRF, we train the NeRFs in Nerfstudio~\cite{nerfstudio} following their paper and render the images according to the camera paths provided for evaluation.

\subsection{Evaluation Metrics}
\paragraph{Optical-Flow-Based Multi-View Consistency.} We follow StyleRF~\cite{liu2023stylerf} and report the optical-flow-based multi-view consistency. Specifically, we employ the optical-flow algorithm RAFT~\cite{flow_Teed_Deng_2020} and softmax splatting~\cite{softmax_splat_Niklaus_Liu_2020} to warp one view to another. We measure the consistency of the stylization using masked RMSE and LPIPS scores~\cite{lpips_Zhang_Isola_Efros_Shechtman_Wang_2018} on the warped and aligned image pairs. Additionally, we calculate short- and long-range consistency scores, comparing adjacent and far-away views, respectively, following~\cite{refart1_Chiang_2021,longshort_Fan_Jiang_Wang_Gong_Xu_Wang_2022,Huang_Tseng_Saini_Singh_Yang_2021}.

\setlength{\tabcolsep}{8pt}
\begin{table}[t]
    \centering
    \begin{tabular}{l|cc|cc}
    \toprule
         Methods&  \multicolumn{2}{c|}{Short-Term}& \multicolumn{2}{c}{Long-Term} \\ \midrule
 & LPIPS& RMSE& LPIPS&RMSE\\ \midrule
         Original & 0.111& 0.025&  0.400&0.064\\
         Instruct-Pix2Pix & 0.374& 0.134& 0.588&0.195\\
         Instruct-NeRF2NeRF & 0.104& 0.023& 0.414&0.061\\
         Ours-Low &   0.086&  0.017& 0.380&0.059\\ 
         Ours & 0.092& 0.019& 0.388&0.062\\ 
    \bottomrule
    \end{tabular}
    \caption{Quantitative results on short-term and long-term multi-view consistency evaluated by LPIPS ($\downarrow$) and RMSE ($\downarrow$).}
    \label{tab:consisentcy}
    \vspace{-3mm}
\end{table}
\setlength{\tabcolsep}{3pt}

\setlength{\tabcolsep}{8pt}
\begin{table}[t]
    \centering
    \begin{tabular}{l|c|c}
    \toprule
         Methods&  BRISQUE ($\downarrow$) & Sharpness ($\uparrow$)\\ \midrule
         Original Scene     &17.46  &  273.28 \\ 
         Instruct-NeRF2NeRF &38.99  & 40.61 \\
         Ours               &17.15  & 255.03 \\
    \bottomrule
    \end{tabular}
    \caption{Quantitative results on the image quality evaluated by BRISQUE and image sharpness.}
    \vspace{-8mm}
    \label{tab:quality}
\end{table}
\setlength{\tabcolsep}{3pt}

\paragraph{Referenceless Image Spatial Quality.} According to our theorem in Formula \ref{eqn:consistency}, blurry smoothed images are easy to get good results under consistency metrics. Therefore, only selecting editing algorithms based on the multi-view consistency may lead to blurry results, which is not optimal. To this end, we incorporate blind image quality measurement using BRISQUE~\cite{brisque} and sharpness score\footnote{\url{https://github.com/opencv/opencv}} to evaluate the quality and sharpness of the images.

\subsection{Per-scene Editing}
We first compare our method with 2D editing methods, i.e., editing the images rendered from the original NeRF~\cite{DBLP:conf/eccv/MildenhallSTBRN20}.
From the results shown in Table~\ref{tab:consisentcy}, the inconsistency between the different views is severe.
The visualization result is shown in the supplementary materials.

By retraining the NeRF with edited images and iteratively updating the training images, Instruct-NeRF2NeRF enforces consistency in the 3D space. 
However, imposing 3D fusion to resolve inconsistency comes at the cost of sacrificing high-resolution details, as demonstrated in Figure~\ref{fig:vis_res_all} and Table~\ref{tab:quality}.
The BRISQUE quality rises from $17.46$ to $38.99$, and the sharpness quality decreases from $273.28$ to $40.61$ compared to the original scene.
In contrast, our method keeps the quality and sharpness of the rendered images almost the same as in the original scene.

We further compare our method to Instruct-NeRF2NeRF on two large-scale scenes, where retaining the fine details of the scene is more difficult.
As visualized in Figure~\ref{fig:vis_res_all}, blurring of the image can be perceived on various views, while the fine details are retained in our method.

\begin{figure}[t]
\centering
\includegraphics[width=0.8\linewidth]{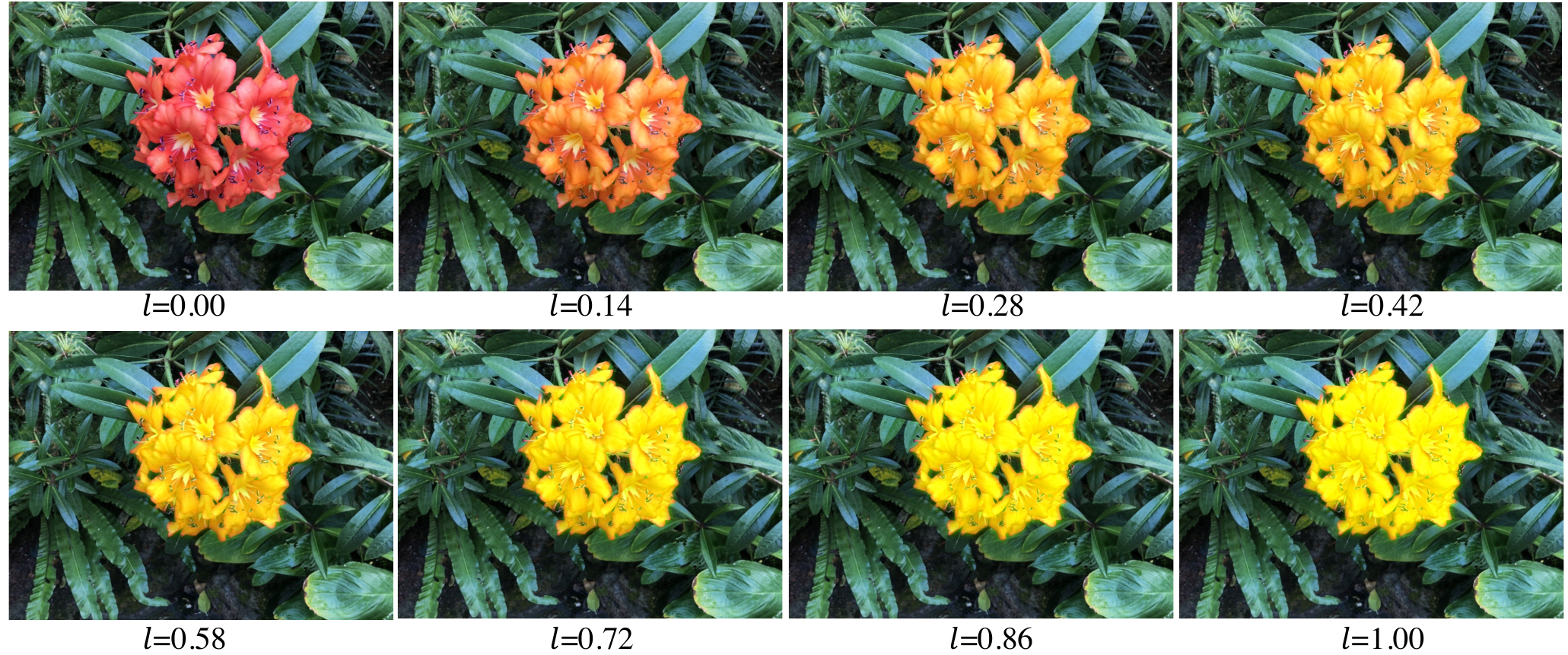}
\caption{Visualization results of controllable style transfer \textit{during network inference}. The style is trained with the instruction \textit{``Turn the flower into yellow"} and adjusted by changing editing level $l$. 
}
\label{fig:vis_control}
\vspace{-3mm}
\end{figure}

\subsection{Intensity Control}
Previous NeRF-editing methods, such as Instruct-NeRF2NeRF, cannot control the level of stylization after training. The only way to adjust the appearance is to regenerate the 2D images of different guidance scales and retrain the original NeRF, which is resource-consuming and has difficulty in fine adjustment.
Benefitting from our feature-space editing regime, the continuous adjustment could be realized by tuning the interpolation indicator $l$.
To show the intensity control, we show the edited scenes with $\l$ rising from $0$ to $1$ in Figure~\ref{fig:vis_control}, from where we see the level of the stylization getting stronger and stronger.

\subsection{Style Transfer}

Previous 3D editing methods usually require training of the editing function for each scene~\cite{haque2023instructnerf2nerf}.
The trained stylization cannot be transferred to other scenarios, which causes high resource consumption when editing a new scene.
In contrast, benefiting from the manipulation of the low-frequency feature space, the trained stylizer in our framework could be directly applied to novel scenes, as shown in Figure~\ref{fig:vis_res_all} lower part.
After training the stylizer of the instruction \textit{``Make it Autumn"} or \textit{``Turn the flowers into yellow"} in the first scene, we directly apply it to the second scene, where the stylizer seamlessly edits this scene to the target style.
This could significantly reduce the workload of editing a 3D scene, and prompt the 3D editing to be applied to a large amount of scenes.
We further train the stylizer in a collection of 2D images, and then transfer it to 3D scenes.

\begin{figure}[t]
\centering
\begin{minipage}[ht]{0.48\linewidth}
\centering
\includegraphics[width=1.\linewidth]{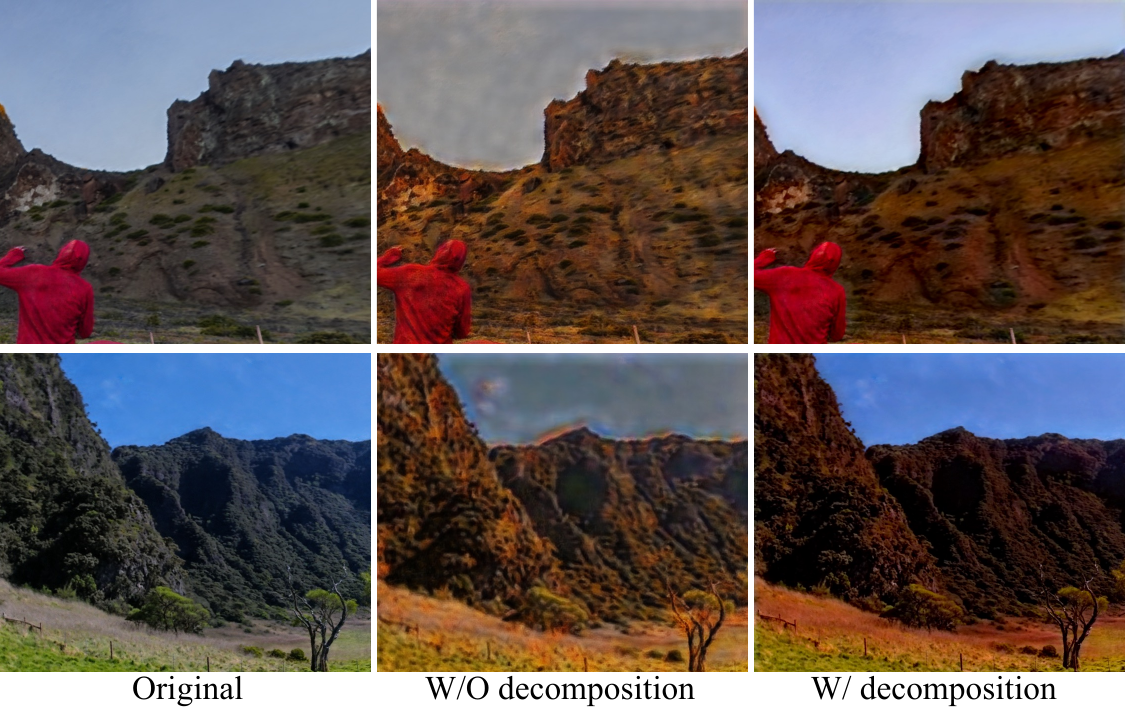}
\caption{Ablation of frequency decomposition. The first row shows the per-scene editing and the second row shows the editing transferred to a novel scene. 
}
\label{fig:ablation_freq_decomp}
\end{minipage}
\hspace{1mm}
\begin{minipage}[ht]{0.48\linewidth}
\centering
\includegraphics[width=1.\linewidth]{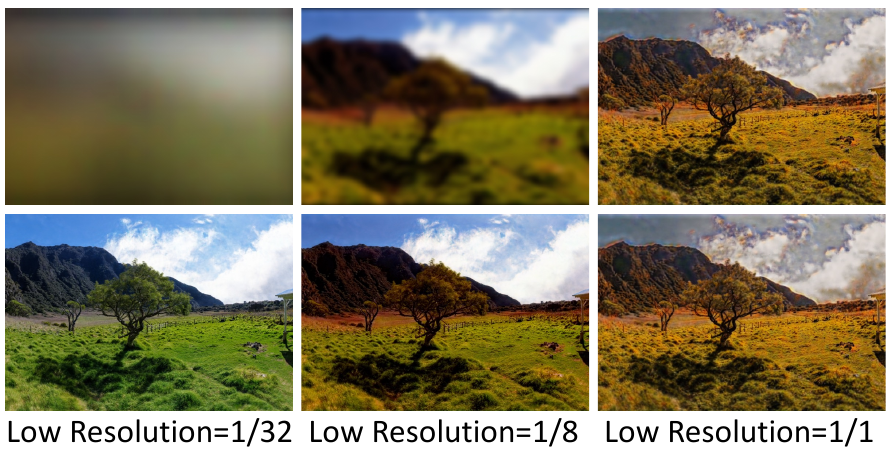}
\caption{
Ablation of different levels of the low-frequency filter. The first row displays the styled low-frequency component and the second row presents the final blended image.
}
\label{fig:ablation_l_level}
\end{minipage}
\end{figure}

\begin{table}[t]
    \centering
    {
    \begin{tabular}{l  c  c c c}
    \toprule
         Methods&  BRISQUE ($\downarrow$) & Sharpness ($\uparrow$) & Long-LPIPS ($\downarrow$) & Long-RMSE ($\downarrow$) \\ 
         \midrule
         W/O decompose     & 20.01  & 193.13 & 0.402  & 0.069 \\ 
         W/ decompose      & 17.15  & 255.03 & 0.388 & 0.062 \\
    \bottomrule
    \end{tabular}}
    \caption{Ablation of the frequency decomposition. The BRISQUE and Sharpness indicating image quality, and the LPIPS and RMSE indicating long-term multi-view consistency are reported.}
    \vspace{-5mm}
    \label{tab:ablation_fd}
\end{table}

\subsection{Ablation Study}

\paragraph{Effect of Frequency Decomposition.} 
To inspect the effect of frequency decomposition, we build a baseline without the decomposition and blending, i.e., there is only feature editing in one branch without the low filter.
The comparison displayed in Figure~\ref{fig:ablation_freq_decomp} shows that editing without decomposition tends to bring artifacts, especially when transferred to a novel scene. Quantitative results in Table \ref{tab:ablation_fd} also show that our full pipeline outperforms baseline without frequency decomposition on image quality and sharpness.
This demonstrates that editing in low-frequency feature space is much easier than in full feature space, which motivates our approach.

\begin{figure}[t]
\centering
\includegraphics[width=0.7\linewidth]{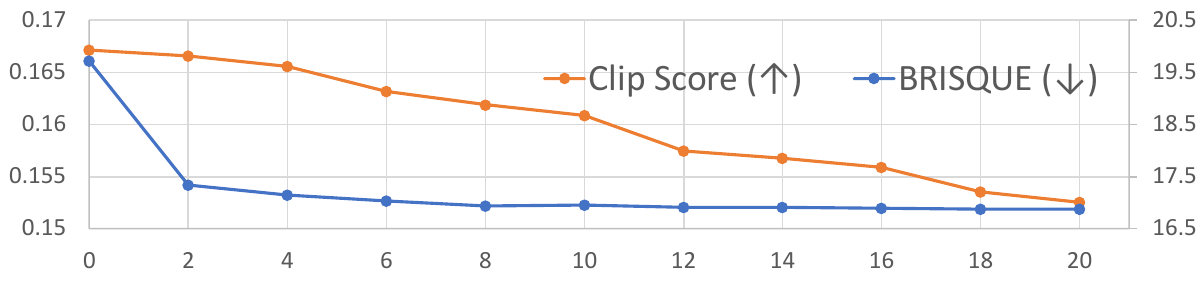}
\caption{Ablation of the low-frequency filter level. We plot the Clip Score indicating the editing effectiveness and the BRISQUE indicating the image quality, w.r.t the downsampling scale of the low-frequency filter. 
The Clip Score is depicted by the orange line and left vertical axis, while the BRISQUE is depicted by the blue line and the right vertical axis.
}
\label{fig:ablat_plot}
\vspace{-5mm}
\end{figure}

\paragraph{Effect of Low-frequency Filter Level.}
The level of the low-frequency space determines where we perform the editing.
Apparently, the frequency could not be unboundedly low, in which case the image contains nearly no information.
Thus, we ablate the level of the low-frequency filter and visualize the results in Figure~\ref{fig:ablation_l_level} and Figure~\ref{fig:ablat_plot}.
We change the downsampling scale of the filter to generate low-frequency images of different resolutions, from which we find that resolutions less than $1/10$ would critically reduce the editing effectiveness, while resolutions greater than $1/4$ would yield artifacts, as illustrated in Figure~\ref{fig:ablat_plot}.
In our experiments, we adopt a downsampling scale of $1/4$ for the balance of effective editing and without artifacts.

\section{Conclusions and Future Work}

This paper studies the problem of high-fidelity and transferable NeRF editing.
The pivotal insight is that the low-frequency components of images, which predominantly define the appearance style, exhibit enhanced multiview consistency after editing compared to their high-frequency counterparts.
This realization has prompted us to decompose the NeRF appearance into low- and high-frequency components and concentrate on editing the feature space of the low-frequency components, resulting in high-fidelity edited scenes. In addition, low-frequency editing in feature space enables stable intensity control and novel scene transfer, which are demonstrated in the experiments.

\paragraph{Limitations and Future Work.}
In some cases, the mixing of the high-frequency component may produce conflicts with the target style, where the details are undesirable in some regions.
In editing these scenarios, an intelligent $\alpha$-blending is needed to bring about a harmonious fusion between the high-frequency details and the edited low-frequency component, which is a future direction.

\appendix
\section{Theoretical Proof of Low-frequency Editing}
\label{Section:Optical:Flow:Consistency}

Consider two images $I_1$ and $I_2$ that describe the 3D scene under different camera poses $\set{C}_1$ and $\set{C}_2$. For simplicity, we consider the setting where both $I_1\in \R^{n\times n}$ and $I_2\in \R^{n\times n}$ are grayscale images and there is a well-defined optical flow $\bs{u}^{\star}:\in R^{2n^2}$ from $I_1$ and $I_2$. For each pixel $\bs{p}\in I_1$, we denote its displacement by $\bs{u}^{\star}(\bs{p}) = (u_x^{\star}(\bs{p}),u_y^{\star}(\bs{p}))$ where $u_x^{\star}(\bs{p})$ and $u_y(\bs{p})$ denote the components in $x$ and $y$, respectively. With this setup, we define the consistency score between $I_1$ and $I_2$ as 
\begin{align}
c(I_1, I_2):= & \sum\limits_{\bs{p}\in I_1}\big(\frac{\partial I_1}{\partial x}(\bs{p})u_x^{\star}(\bs{p}) + \frac{\partial I_1}{\partial y}(\bs{p})u_y^{\star}(\bs{p}) \\
& \qquad + I_2(\bs{p})-I_1(\bs{p})\big)^2 
\label{Eq:Based:Consistency}
\end{align}
where $\frac{\partial I_1}{\partial x}(\bs{p})u_x^{\star}(\bs{p}) + \frac{\partial I_1}{\partial y}(\bs{p})u_y^{\star}(\bs{p})+ I_2(p) - I_1(\bs{p}) = 0$ is the first order approximation of the color consistency constraint $I_1(\bs{p}) = I_2(\bs{p}+\bs{u}^{\star}(\bs{p}))$.

The optimal flow in $\bs{u}^{\star}$ is defined by minimizing $c(I_1,I_2)$ and a regularization term that enforces $\bs{u}$ is smooth:
\begin{align}
& \bs{u}^{\star} = \underset{\bs{u}}{\textup{argmin}} \quad \mu\sum\limits_{\bs{p}\in I_1}\big(\frac{\partial I_1}{\partial x}(\bs{p})u_x^{\star}(\bs{p}) + \frac{\partial I_1}{\partial y}(\bs{p})u_y^{\star}(\bs{p})\nonumber \\
& \quad +(I_2(\bs{p})-I_1(\bs{p}))\big)^2+  \sum\limits_{(\bs{p}_1,\bs{p}_2)\in \set{E}}\|u(\bs{p}_1)-u(\bs{p}_2)\|^2
\end{align} 
where $\set{E}$ collects adjacent pixel pairs. The trade-off parameter $\mu$ is a small value that prioritizes that the optical flow is smooth.

Introduce $A(I_1) = \diag(A(\bs{p})\in \R^{2n\times n}$ and $\bs{b}(I_1,I_2) = (\bs{b}(\bs{p}))\in \R^{n}$, where
\begin{align*}
A(\bs{p}) & = \left(
\begin{array}{c}
\big(\frac{\partial I_1}{\partial x})(\bs{p})\\
\big(\frac{\partial I_1}{\partial y})(\bs{p})
\end{array}
\right)\\
\bs{b}(\bs{p}) & =  (I_2(\bs{p})-I_1(\bs{p})).
\end{align*}
Let $L\in \R^{n^2\times n^2}$ be the Laplacian matrix of $\set{E}$. It is clear that the optimal solution
\begin{align}
\bs{u}^{\star} & =\underset{\bs{u}}{\textup{argmin}} \ \mu \|A(I_1)^T\bs{u}+\bs{b}(I_1,I_2)\|^2 + \bs{u}^T (L\otimes I_2)\bs{u} \nonumber \\
& = -\mu \Big(\mu A(I_1)A(I_1)^T+(L\otimes I_2)\Big)^{-1}A(I_1)\bs{b}(I_1,I_2)
\label{Eq:u:opt}
\end{align}

Substituting (\ref{Eq:u:opt}) into (\ref{Eq:Based:Consistency}), we arrive at the following definition 
\begin{align}
c(I_1, I_2)  &=\|\big(I_n-\mu A(I_1)^T(\mu A(I_1)A(I_1)^T+ (L\otimes I_2))^{-1} \nonumber \\
& \qquad A(I_1)\big)\bs{b}(I_1,I_2)\|^2 \label{Eq:Consistency:Definition} 
\end{align}

Denote $\set{S}: \R^{n\times n}\rightarrow \R^{n\times n}$ as the low-pass filter that performs image smoothing. Let $\set{T}:\R^{n\times n} \rightarrow \R^{n\times n}$ be the stylization operation. Without losing generality, we assume that 
$$
\set{T}\circ \set{S} \approx \set{S}\circ \set{T},
$$
which means that we get similar results when stylizing the smoothed image and smoothing the stylized image. This assumption is applicable to most stylization and smoothing operations.

In the following, we present our main theoretical result.
\begin{theorem}
Suppose there is color consistency between $I_1$ and $I_2$.
$$
c\big(\set{T}(\set{S}(I_1)), \set{T}(\set{S}(I_2))\big) < c\big(\set{T}(I_1), \set{T}(I_2)\big).
$$
\end{theorem}
In other words, if we do stylization after smoothing. View consistency is superior to that without smoothing.

\noindent\textsl{Proof:} We show the following theorem, whose proof is deferred to Section~\ref{Proof:Lemma:Smoothing}. 

\begin{theorem}
Suppose there is color consistency between $I_1$ and $I_2$.  Given a smoothing filter $\set{S}$, we have
\begin{equation}
c(\set{S}(I_1),\set{S}(I_2)) < c(I_1, I_2)    
\end{equation}
\label{Lemma:Smoothing}
\end{theorem}

Since $\set{T}\cdot \set{S} \approx \set{S} \cdot \set{T} $. We have 
\begin{align*}
&\ \ c((\set{T}\circ \set{S})(I_1),(\set{T}\circ \set{S})(I_2)) \\
\approx  & \ \ c((\set{S}\circ \set{T})(I_1),(\set{S}\circ \set{T})(I_2))\\
< & \ \ c (\set{T}(I_1), \set{T}(I_2))
\end{align*}
\hfill $\square$

\subsection{Proof of Theorem~\ref{Lemma:Smoothing}}
\label{Proof:Lemma:Smoothing}

Note that, due to the color consistency assumption, we have
\begin{equation}
\bs{1}^T\bs{b}(I_1, I_2) = 0.
\label{Eq:Color:Consistency}
\end{equation}

Denote $I_1^s = \set{S}(I_1)$ and $I_2^s = \set{S}(I_2)$. Let $A = A(I_1)$ and $A^s = A(I_1^s)$
It is sufficient to show that
\begin{align}
\|(I_n-D(A_s))\bs{b}(I_1^s,I_2^s)\|<  \|(I_n-D(A))\bs{b}(I_1,I_2)\| 
\label{Eq:2}
\end{align}
where
$$
D(A) = \mu A^T\big(\mu AA^T + (L\otimes I_2)\big)^{-1}A.
$$

Let $S \in \R^{n^2\times n^2}$ be the matrix representation of the smoothing operator $\set{S}$. It is clear that all elements of $S$ are non-negative. Furthermore, as a smoothing operator, we have
$$
S\bs{1} = \bs{1}, \qquad S^T\bs{1}=\bs{1},
$$ 
i.e., the row and column sums are $1$. 
Using this notation, it is easy to see that 
\begin{equation}
A_s = (S\otimes I_2)A, \quad \bs{b}(I_1^s,I_2^s) = S\big(\bs{b}(I_1,I_2)\big)
\label{Eq:Smoothing:Operator}
\end{equation}
In the following, we show that for small $\mu$, 
$\forall \bs{v}\in \R^n, \bs{1}^T\bs{v} =0 $,
\begin{equation}
\|(I_n-D(A_s))S\bs{v}\|^2<  \|(I_n-D(A))\bs{v}\|^2  
\label{Eq:A:3}
\end{equation}

To this end, we first derive a closed-form expression of $D(A)$.Let $\sigma_i$ and $\bs{u}_i$ collect the eigenvalues and eigenvectors of $L$, where $\sigma_1 = 0$. Denote $U = (\bs{u}_2,\cdots, \bs{u}_n)$ and $\Sigma = \textup{diag}(\lambda_i)$. Introduce
\begin{align*}
\overline{U} = A^T (U\otimes I_2), &\quad  \overline{E} = A^T (\bs{1}\otimes I_2),\\
\overline{U}_s = A^T(S^TU\otimes I_2), & \quad \overline{E}_s = A^T (S^T\bs{1}\otimes I_2).
\end{align*}
As $S^T\bs{1} = \bs{1}$. We have $\overline{E}_s = \overline{E}$.
The following lemma, whose proof is deferred to Section~\ref{Lem:D:Expression:Proof}, provides an expression of $D(A)$.

\begin{lem}
In the small $\mu$ regime, we have
\begin{align}
D(A) & = \overline{E}({\overline{E}}^T\overline{E})^{-1}\overline{E}^T + \mu G
\label{Eq:D:Expression}
\end{align}
where
$$
G = \Big(\overline{L}^{+} -\overline{E}({\overline{E}}^T\overline{E})^{-1}\overline{E}^T\overline{L}^{+}-\overline{L}^{+}\overline{E}({\overline{E}}^T\overline{E})^{-1}\overline{E}^T\Big)
$$
where $\overline{L}^{+} = \overline{U}(\Sigma\otimes I_2)\overline{U}^T$.
\label{Lem:D:Expression}
\end{lem}

Substituting (\ref{Eq:D:Expression}) into~\ref{Eq:A:3}, in the small $\mu$ regime, it is sufficient to show that,
$$
\|(I_n -  \overline{E}({\overline{E}}^T\overline{E})^{-1}\overline{E}^T)S\bs{v}\|^2 \leq  \|(I_n -  \overline{E}({\overline{E}}^T\overline{E})^{-1}\overline{E}^T)\bs{v}\|^2
$$
which is equivalent to 
\begin{align*}
\bs{v}^TS^2 \bs{v} < \bs{v}^T\bs{v}.
\end{align*}
As $\overline{E}^T\bs{v} = \overline{E}^T\bs{v} = 0$.

Note that
\begin{align*}
& \bs{v}^T\bs{v} -\bs{v}^TS^S\bs{v} \\ 
= & \frac{1}{2}\sum\limits_{i=1}^{n^2}\big(\sum\limits_{j=1}^{n^2}S_{ij}v_j^2 + \sum\limits_{k=1}^{n^2}S_{ik}v_k^2\big) - \bs{v}^TS^S\bs{v} \\
= & \frac{1}{2}\sum\limits_{i=1}^{n^2}\sum\limits_{j=1}^{n^2}\sum\limits_{k=1}^{n^2}S_{ij}S_{ik}(v_j^2+v_k^2) - \bs{v}^TS^S\bs{v} \\
= & \frac{1}{2} \sum\limits_{i=1}^{n^2}\sum\limits_{j=1}^{n^2}\sum\limits_{k=1}^{n^2}S_{ij}S_{ik}(v_j-v_k)^2 > 0
\end{align*}
due to $\bs{1}^T\bs{v} = 0$.
\hfill $\square$

\subsubsection{Proof of Lemma~\ref{Lem:D:Expression}}
\label{Lem:D:Expression:Proof}

Note that
\begin{align*}
& \mu AA^{T} + \overline{L} \nonumber \\
= & (U, E)
\left(
\begin{array}{cc}
\Sigma +  \mu {\overline{U}}^T\overline{U} &  \mu {\overline{U}}^T\overline{E} \\
 \mu {\overline{E}}^T\overline{U} & \mu {\overline{E}}^T\overline{E}
\end{array}
\right)(U, E)^T,
\end{align*}

Therefore,
\begin{align*}
& A^T\big(\mu AA^{T} +  \overline{L}\big)^{-1}A \nonumber \\
= &  (\overline{U}, \overline{E})
\left(
\begin{array}{cc}
\Sigma +  \mu {\overline{U}}^T\overline{U} &  \mu {\overline{U}}^T\overline{E} \\
 \mu {\overline{E}}^T\overline{U} & \mu {\overline{E}}^T\overline{E}
\end{array}
\right)^{-1}(\overline{U}, \overline{E})^T.
\end{align*}
Applying Schur complement, we have
\begin{equation}
 \left(
\begin{array}{cc}
\Sigma +  \mu {\overline{U}}^T\overline{U} &  \mu {\overline{U}}^T\overline{E} \\
 \mu {\overline{E}}^T\overline{U} & \mu {\overline{E}}^T\overline{E}
\end{array}
\right)^{-1} =  \left(
\begin{array}{cc}
M_{11} & M_{12} \\
M_{12}^T & M_{22}
\end{array}
\right)
\label{Eq:Schur:Complement}
\end{equation}
where
\begin{align}
M_{22}= & \Big(\mu {\overline{E}}^T\overline{E} - \mu^2 {\overline{E}}^T\overline{U}\big(\Sigma +  \mu {\overline{U}}^T\overline{U}\big)^{-1}{\overline{U}}^T\overline{E}\Big)^{-1}, \nonumber \\
M_{12}= &  -\mu \big(\Sigma +  \mu {\overline{U}}^T\overline{U}\big)^{-1}{\overline{U}}^T\overline{E}M_{22},\nonumber \\
M_{11}= &\big(\Sigma +  \mu {\overline{U}}^T\overline{U}\big)^{-1} + \mu^2 \big(\Sigma +  \mu {\overline{U}}^T\overline{U}\big)^{-1}{\overline{U}}^T\overline{E}\nonumber \\
& M_{22}{\overline{E}}^T\overline{U}\big(\Sigma +  \mu {\overline{U}}^T\overline{U}\big)^{-1}.\nonumber 
\end{align}
Omitting high-order terms of $\mu$ in $M_{11}$, $M_{12}$, and $M_{22}$, we have
\begin{align}
 M_{22}= & \frac{1}{\mu} ({\overline{E}}^T\overline{E})^{-1} \label{Eq:M22:Approx}\\ 
 M_{12}= &-\Sigma^{-1}{\overline{U}}^T\overline{E}({\overline{E}}^T\overline{E} )^{-1}\label{Eq:M12:Approx}\\  
 M_{11}= &\Sigma^{-1}\label{Eq:M11:Approx}   
\end{align}
Note that
\begin{align}
&  \mu A^T\big(\mu AA^{T} +  \overline{L}\big)^{-1}A \nonumber \\
= &\mu (\overline{U},\overline{E})\left(
\begin{array}{cc}
M_{11} &  M_{12} \\
M_{12}^T &  M_{22}
\end{array}
\right)(\overline{U},\overline{E})^T \nonumber \\
= & \mu\Big(\overline{U}M_{11}\overline{U}^T + \overline{U}M_{12}\overline{E}^T + \overline{E}M_{12}^T\overline{U}^T+\overline{E}M_{22}\overline{E}^T\Big).
\label{Eq:A:2:1}
\end{align}
Substituting (\ref{Eq:M22:Approx}), (\ref{Eq:M12:Approx}), and (\ref{Eq:M11:Approx}) into (\ref{Eq:A:2:1}),  we have
\begin{align}
D(A) = & \mu A^T\big(\mu AA^{T} +  \overline{L}\big)^{-1}A \nonumber \\
=& \overline{E}({\overline{E}}^T\overline{E})^{-1}\overline{E}^T + \mu \Big(\overline{U}\Sigma^{-1}\overline{U}^T -\overline{E}({\overline{E}}^T\overline{E})^{-1}\overline{E}^T\nonumber \\
& \overline{U}\Sigma^{-1}\overline{U}^T-\overline{U}\Sigma^{-1}\overline{U}^T\overline{E}({\overline{E}}^T\overline{E})^{-1}\overline{E}^T\Big)
\end{align}
\hfill $\square$
\section{Implementation Details}

\noindent{\textbf{Network Architecture.}} \textit{Feature-based framework}. The backbone architecture of NeRF is the `nerfacto' from NeRFStudio~\cite{nerfstudio}. For the high-resolution branch that reconstructs the original scene, we set the size of the hashmap for the base MLPs to $2^{24}$ and the maximum resolution of the hashmap to $20480$. The feature field consists of another hashmap of size $2^{23}$ and $4$ layers MLPs that rendered $256$ channels of distilled VGG~\cite{VGG} feature. The low-frequency filter consists of $3$ convolution layers with kernel size $5$ and ReLU~\cite{ReLU} activation. The filtered feature map is downscaled to $1/4$ resolution of the original feature map by bilinear interpolation and fed into the style module. The style module consists of $2$ convolution layers with Instance Normalization~\cite{instnorm} and ReLU activation, followed by $5$ residual blocks~\cite{resnet}. The decoder that decodes the image from (styled) feature maps consists of $5$ residual blocks and $3$ convolution layers. For each residual block in our framework, the batch normalization layers~\cite{BN} are removed for better multi-view consistency. \textit{Per-scene high-fidelity enhancing framework.} Our frequency decomposition mechanism can also be applied to enhance the per-scene editing results built on Instrcut-NeRF2NeRF~\cite{haque2023instructnerf2nerf}. Specifically, we regard the original Instruct-NeRF2NeRF framework as the low-frequency branch. And add another `nerfacto' branch to reconstruct the high-fidelity original scene. We then extract the high-frequency details $I_{oHigh}$ and low-frequency image $I_{oLow}$ from the high-fidelity original scene by Gaussian low-pass filters. Similarly, we extract $I_{eHigh}$ and $I_{eLow}$ from the Instruct-NeRF2NeRF branch. The final enhanced result $I_{en}$ can then be obtained by $I_{en}=I_{eLow} + I_{oHigh}$. For edits with geometry changes, directly blending high-frequency details into the edited scene may produce artifacts in the deformed regions. We can compute an alpha mask $\alpha$ based on the difference between the edited and original low-frequency images for blending as $\alpha=\text{abs}(I_{eLow}-I_{oLow})$, where abs is the function to calculate the absolute values. The enhanced edited image $I_{en}$ can then be obtained by $I_{en}=I_{eLow}+ \lambda_1 \alpha I_{eHigh} + \lambda_2 (1-\alpha ) I_{oHigh}$, with $\lambda_{1}, \lambda_{2} \in [0, 1]$ for further strength adjustment.  

\noindent{\textbf{Training Setting.}} The `nerfacto' that reconstructs the original scene is trained for $100,000$ iterations with mean square error (MSE) loss and Adam~\cite{Adam} optimizer. The learning rate starts from $0.01$ and gradually degrades to $0.0001$. The low-frequency branch is trained within a two-stage manner as demonstrated in Section 3.5 of our main paper. In the first stage, the feature field, low-frequency filter, and decoder are trained for $40,000$ iterations with Adam optimizer and learning rate $0.0001$. In the second stage, we train the style module for $40,000$ iterations with Adam optimizer and learning rate $0.0001$. The setup of the iterative dataset update is the same as Instruct-NeRF2NeRF~\cite{haque2023instructnerf2nerf}. Concurrent works \cite{ED_NERF,DreamEditor,GaussianEditor} utilize 2D~\cite{segmentanything} or 3D masks to enhance the local editing results. Following these techniques, we also utilize 2D masks to enhance the 2D edited results from Instruct-Pix2Pix~\cite{brooks2022instructpix2pix}. Specifically, the enhanced edited image $I_{en}$ is recomposed by $I_{en} = M\cdot I_{ed} + (1-M)\cdot I_{or}$, where $M$ indicates the segmentation mask, $I_{ed}$ indicates the edited image and $I_{or}$ indicates the original image.

\section{More Visualization Results}
\begin{figure*}[!h]
\centering
\vspace{0.8in}
\includegraphics[width=1.0\linewidth]{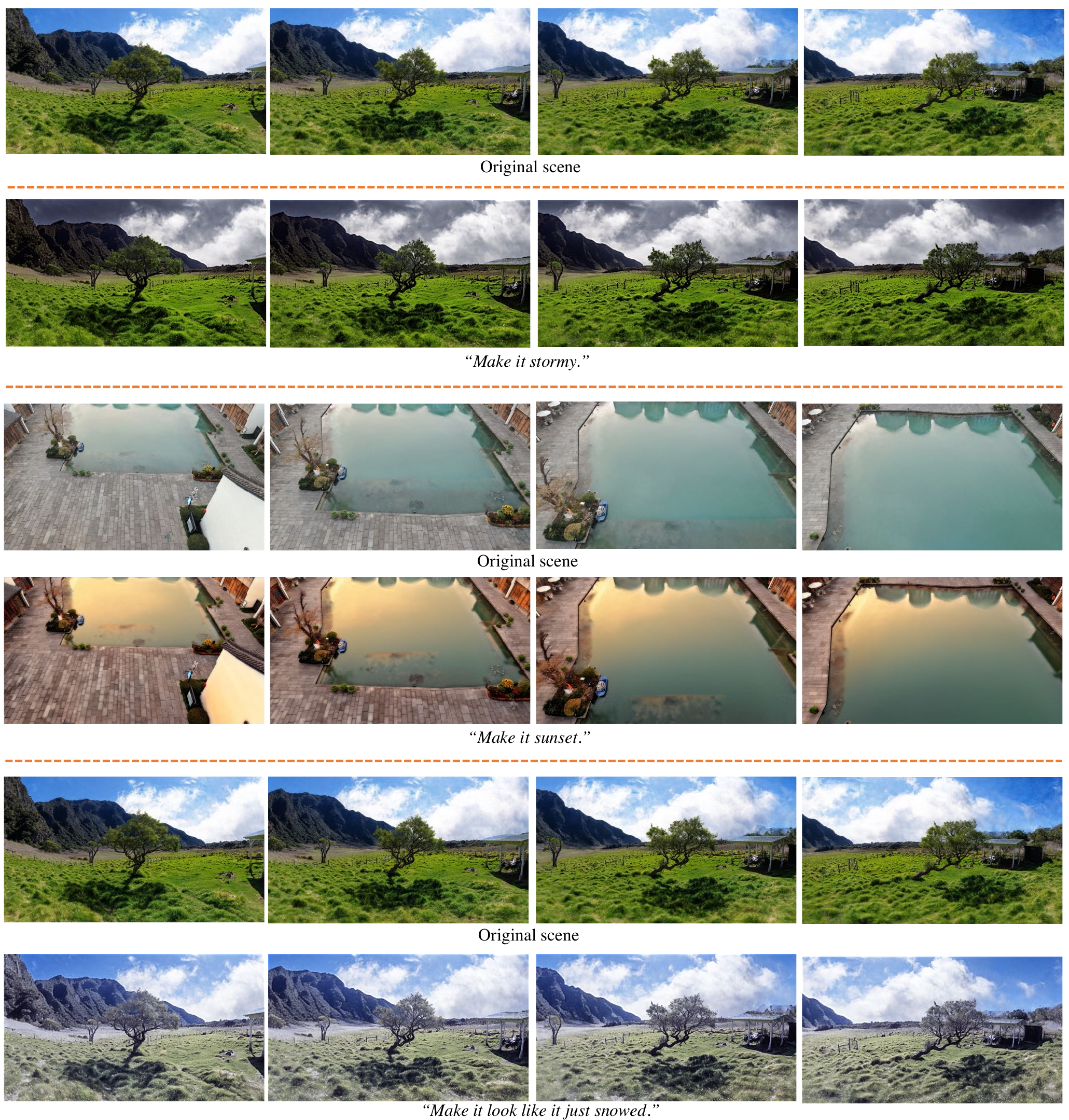}
\caption{\textbf{Visualization of our edited 3D scenes}. Our method can perform global editing on 3D scenes without losing high-fidelity details.}
\label{fig:more_vis0}
\end{figure*}

\begin{figure*}[!h]
\centering
\vspace{0.8in}
\includegraphics[width=1.0\linewidth]{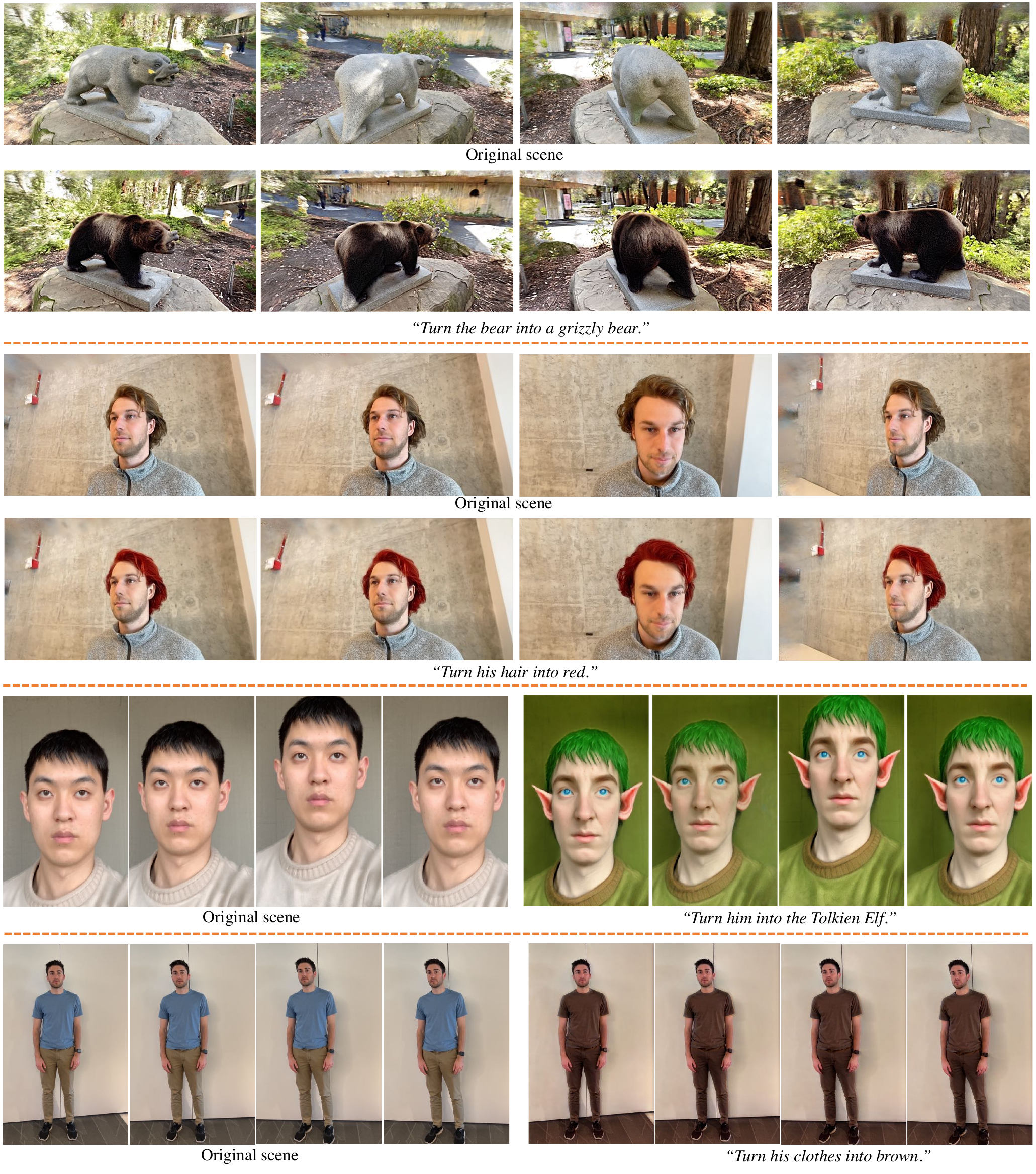}
\caption{\textbf{Visualization of our edited 3D scenes}. Our method is able to edit specific objects or specific parts of a 3D scene without losing high-fidelity details.}
\label{fig:more_vis1}
\end{figure*}

\begin{figure*}[!h]
\centering
\vspace{0.8in}
\includegraphics[width=1.0\linewidth]{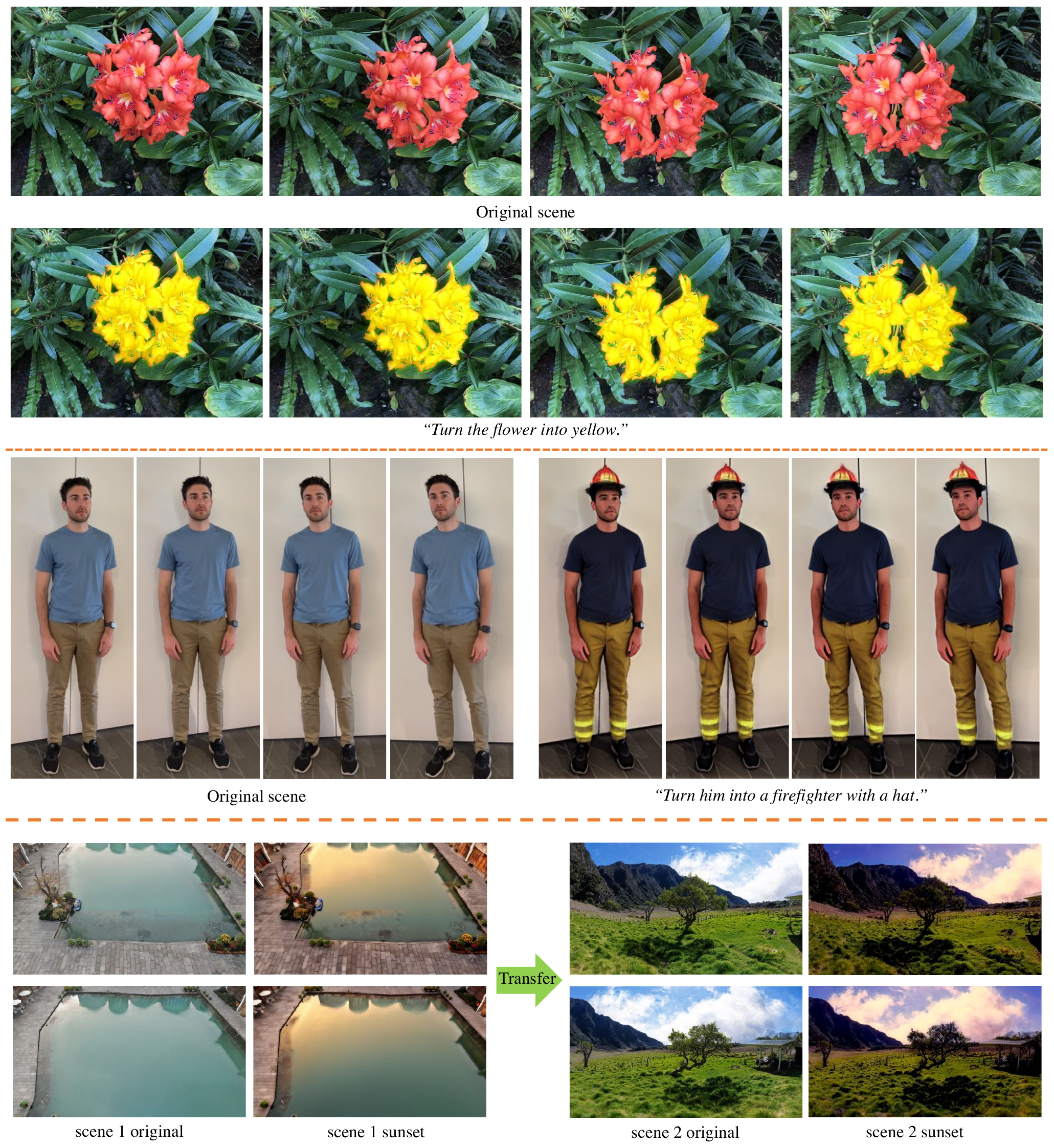}
\caption{\textbf{Visualization of our edited 3D scenes}. Results on more datasets and results of more style transfers are displayed.}
\label{fig:more_vis2}
\end{figure*}

\begin{figure*}[!h]
\centering
\vspace{1in}
\includegraphics[width=1.0\linewidth]{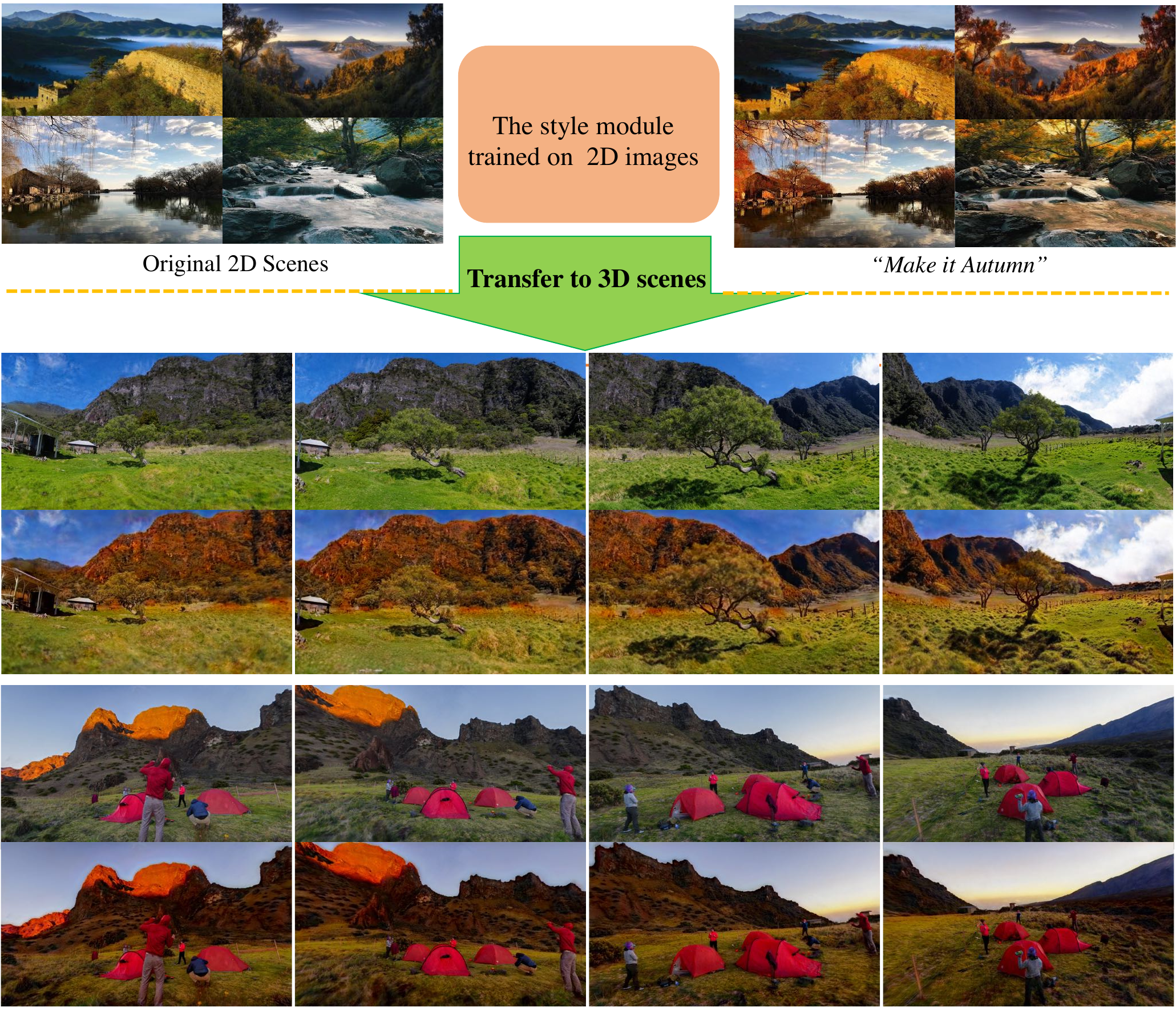}
\caption{\textbf{Visualization of our transferable style editing}. Our style modules, initially trained on a collection of 2D images, are able to transfer to 3D scenes for multi-view consistent editing \textit{without the necessity for retraining}.}
\label{fig:more_web_trans}
\end{figure*}

\begin{figure*}[!h]
\centering
\includegraphics[width=1.0\linewidth]{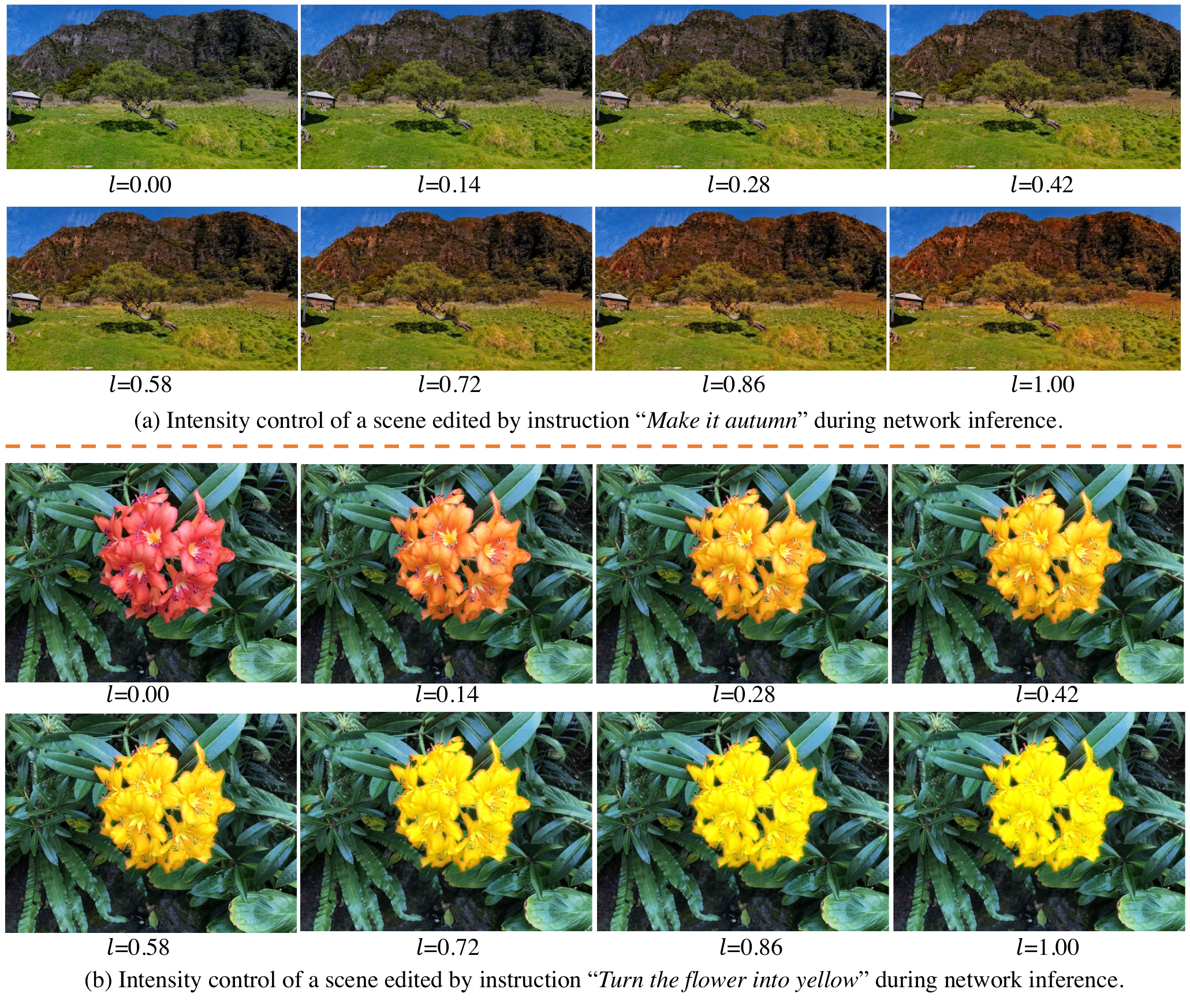}
\caption{\textbf{Visualization of intensity control}. Our framework allows easy adjustment of intensity levels within edited 3D scenes, which is achievable \textit{during network inference} for dynamic control. Figure (a) illustrates the precise intensity control of local edits applied to a specific object, while Figure (b) demonstrates the application of global edits across the entire scene.}
\label{fig:alpha_control}
\end{figure*}

We show more visualization of edited 3D scenes in Figure \ref{fig:more_vis0}, \ref{fig:more_vis1}, and \ref{fig:more_vis2}. Our frequency decomposed editing network can produce various high-fidelity edited 3D scenes with instructions. In Figure \ref{fig:more_web_trans}, we show that the style module trained with some 2D images can be directly transferred to various 3D scenes \textit{without retraining}. In Figure \ref{fig:alpha_control}, we show more results of intensity control \textit{during network inference}.


\newpage

\newpage

\newpage

\newpage

\newpage

%
%
\bibliographystyle{splncs04}
\bibliography{main}

\begin{thebibliography}{10}
\providecommand{\url}[1]{\texttt{#1}}
\providecommand{\urlprefix}{URL }
\providecommand{\doi}[1]{https://doi.org/#1}

\bibitem{barron2022mipnerf360}
Barron, J.T., Mildenhall, B., Verbin, D., Srinivasan, P.P., Hedman, P.: Mip-nerf 360: Unbounded anti-aliased neural radiance fields. CVPR  (2022)

\bibitem{light2_Boss_Jampani_Braun_Liu}
Boss, M., Jampani, V., Braun, R., Liu, C., Barron, J.T., Lensch, H.P.A.: Neural-pil: Neural pre-integrated lighting for reflectance decomposition. In: Ranzato, M., Beygelzimer, A., Dauphin, Y.N., Liang, P., Vaughan, J.W. (eds.) Advances in Neural Information Processing Systems 34: Annual Conference on Neural Information Processing Systems 2021, NeurIPS 2021, December 6-14, 2021, virtual. pp. 10691--10704 (2021), \url{https://proceedings.neurips.cc/paper/2021/hash/58ae749f25eded36f486bc85feb3f0ab-Abstract.html}

\bibitem{brooks2022instructpix2pix}
Brooks, T., Holynski, A., Efros, A.A.: Instructpix2pix: Learning to follow image editing instructions. In: {IEEE/CVF} Conference on Computer Vision and Pattern Recognition, {CVPR} 2023, Vancouver, BC, Canada, June 17-24, 2023. pp. 18392--18402. {IEEE} (2023). \doi{10.1109/CVPR52729.2023.01764}, \url{https://doi.org/10.1109/CVPR52729.2023.01764}

\bibitem{DINO_Caron_Touvron_Misra_Jegou_Mairal_Bojanowski_Joulin_2021}
Caron, M., Touvron, H., Misra, I., Jegou, H., Mairal, J., Bojanowski, P., Joulin, A.: Emerging properties in self-supervised vision transformers. In: 2021 IEEE/CVF International Conference on Computer Vision (ICCV) (Oct 2021). \doi{10.1109/iccv48922.2021.00951}, \url{http://dx.doi.org/10.1109/iccv48922.2021.00951}

\bibitem{GaussianEditor}
Chen, Y., Chen, Z., Zhang, C., Wang, F., Yang, X., Wang, Y., Cai, Z., Yang, L., Liu, H., Lin, G.: Gaussianeditor: Swift and controllable 3d editing with gaussian splatting. CoRR  \textbf{abs/2311.14521} (2023). \doi{10.48550/ARXIV.2311.14521}, \url{https://doi.org/10.48550/arXiv.2311.14521}

\bibitem{refart1_Chiang_2021}
Chiang, P., Tsai, M., Tseng, H., Lai, W., Chiu, W.: Stylizing 3d scene via implicit representation and hypernetwork. In: {IEEE/CVF} Winter Conference on Applications of Computer Vision, {WACV} 2022, Waikoloa, HI, USA, January 3-8, 2022. pp. 215--224. {IEEE} (2022). \doi{10.1109/WACV51458.2022.00029}, \url{https://doi.org/10.1109/WACV51458.2022.00029}

\bibitem{paintnesf}
Duan, H., Wang, M., Li, Y., Yang, Y.: Paintnesf: Artistic creation of stylized scenes with vectorized 3d strokes. CoRR  \textbf{abs/2311.15637} (2023). \doi{10.48550/ARXIV.2311.15637}, \url{https://doi.org/10.48550/arXiv.2311.15637}

\bibitem{longshort_Fan_Jiang_Wang_Gong_Xu_Wang_2022}
Fan, Z., Jiang, Y., Wang, P., Gong, X., Xu, D., Wang, Z.: Unified implicit neural stylization. In: Avidan, S., Brostow, G.J., Ciss{\'{e}}, M., Farinella, G.M., Hassner, T. (eds.) Computer Vision - {ECCV} 2022 - 17th European Conference, Tel Aviv, Israel, October 23-27, 2022, Proceedings, Part {XV}. Lecture Notes in Computer Science, vol. 13675, pp. 636--654. Springer (2022). \doi{10.1007/978-3-031-19784-0\_37}, \url{https://doi.org/10.1007/978-3-031-19784-0\_37}

\bibitem{GaussianEditor2}
Fang, J., Wang, J., Zhang, X., Xie, L., Tian, Q.: Gaussianeditor: Editing 3d gaussians delicately with text instructions. CoRR  \textbf{abs/2311.16037} (2023). \doi{10.48550/ARXIV.2311.16037}, \url{https://doi.org/10.48550/arXiv.2311.16037}

\bibitem{PIE-NeRF}
Feng, Y., Shang, Y., Li, X., Shao, T., Jiang, C., Yang, Y.: Pie-nerf: Physics-based interactive elastodynamics with nerf. CoRR  \textbf{abs/2311.13099} (2023). \doi{10.48550/ARXIV.2311.13099}, \url{https://doi.org/10.48550/arXiv.2311.13099}

\bibitem{img2_Gatys_Ecker_Bethge_2016}
Gatys, L.A., Ecker, A.S., Bethge, M.: Image style transfer using convolutional neural networks. In: 2016 IEEE Conference on Computer Vision and Pattern Recognition (CVPR) (Jun 2016). \doi{10.1109/cvpr.2016.265}, \url{http://dx.doi.org/10.1109/cvpr.2016.265}

\bibitem{ReLU}
Glorot, X., Bordes, A., Bengio, Y.: Deep sparse rectifier neural networks. In: Gordon, G.J., Dunson, D.B., Dud{\'{\i}}k, M. (eds.) Proceedings of the Fourteenth International Conference on Artificial Intelligence and Statistics, {AISTATS} 2011, Fort Lauderdale, USA, April 11-13, 2011. {JMLR} Proceedings, vol.~15, pp. 315--323. JMLR.org (2011), \url{http://proceedings.mlr.press/v15/glorot11a/glorot11a.pdf}

\bibitem{haque2023instructnerf2nerf}
Haque, A., Tancik, M., Efros, A., Holynski, A., Kanazawa, A.: Instruct-nerf2nerf: Editing 3d scenes with instructions. In: Proceedings of the IEEE/CVF International Conference on Computer Vision (2023)

\bibitem{resnet}
He, K., Zhang, X., Ren, S., Sun, J.: Deep residual learning for image recognition. In: 2016 {IEEE} Conference on Computer Vision and Pattern Recognition, {CVPR} 2016, Las Vegas, NV, USA, June 27-30, 2016. pp. 770--778. {IEEE} Computer Society (2016). \doi{10.1109/CVPR.2016.90}, \url{https://doi.org/10.1109/CVPR.2016.90}

\bibitem{imgart1_Hertzmann_1998}
Hertzmann, A.: Painterly rendering with curved brush strokes of multiple sizes. In: Proceedings of the 25th annual conference on Computer graphics and interactive techniques - SIGGRAPH ’98 (Jan 1998). \doi{10.1145/280814.280951}, \url{http://dx.doi.org/10.1145/280814.280951}

\bibitem{Huang_Tseng_Saini_Singh_Yang_2021}
Huang, H.P., Tseng, H.Y., Saini, S., Singh, M., Yang, M.H.: Learning to stylize novel views. International Conference on Computer Vision,International Conference on Computer Vision  (Jan 2021)

\bibitem{Huang_Belongie_2017}
Huang, X., Belongie, S.: Arbitrary style transfer in real-time with adaptive instance normalization. In: 2017 IEEE International Conference on Computer Vision (ICCV) (Oct 2017). \doi{10.1109/iccv.2017.167}, \url{http://dx.doi.org/10.1109/iccv.2017.167}

\bibitem{refart3_Huang_He_Yuan_Lai_Gao}
Huang, Y., He, Y., Yuan, Y., Lai, Y., Gao, L.: Stylizednerf: Consistent 3d scene stylization as stylized nerf via 2d-3d mutual learning. In: {IEEE/CVF} Conference on Computer Vision and Pattern Recognition, {CVPR} 2022, New Orleans, LA, USA, June 18-24, 2022. pp. 18321--18331. {IEEE} (2022). \doi{10.1109/CVPR52688.2022.01780}, \url{https://doi.org/10.1109/CVPR52688.2022.01780}

\bibitem{FaceCLIPNeRF}
Hwang, S., Hyung, J., Kim, D., Kim, M., Choo, J.: Faceclipnerf: Text-driven 3d face manipulation using deformable neural radiance fields. In: {IEEE/CVF} International Conference on Computer Vision, {ICCV} 2023, Paris, France, October 1-6, 2023. pp. 3446--3456. {IEEE} (2023). \doi{10.1109/ICCV51070.2023.00321}, \url{https://doi.org/10.1109/ICCV51070.2023.00321}

\bibitem{BN}
Ioffe, S., Szegedy, C.: Batch normalization: Accelerating deep network training by reducing internal covariate shift. In: Bach, F.R., Blei, D.M. (eds.) Proceedings of the 32nd International Conference on Machine Learning, {ICML} 2015, Lille, France, 6-11 July 2015. {JMLR} Workshop and Conference Proceedings, vol.~37, pp. 448--456. JMLR.org (2015), \url{http://proceedings.mlr.press/v37/ioffe15.html}

\bibitem{Johnson_Alahi_Fei-Fei_2016}
Johnson, J., Alahi, A., Fei{-}Fei, L.: Perceptual losses for real-time style transfer and super-resolution. In: Leibe, B., Matas, J., Sebe, N., Welling, M. (eds.) Computer Vision - {ECCV} 2016 - 14th European Conference, Amsterdam, The Netherlands, October 11-14, 2016, Proceedings, Part {II}. Lecture Notes in Computer Science, vol.~9906, pp. 694--711. Springer (2016). \doi{10.1007/978-3-319-46475-6\_43}, \url{https://doi.org/10.1007/978-3-319-46475-6\_43}

\bibitem{Adam}
Kingma, D.P., Ba, J.: Adam: {A} method for stochastic optimization. In: Bengio, Y., LeCun, Y. (eds.) 3rd International Conference on Learning Representations, {ICLR} 2015, San Diego, CA, USA, May 7-9, 2015, Conference Track Proceedings (2015), \url{http://arxiv.org/abs/1412.6980}

\bibitem{segmentanything}
Kirillov, A., Mintun, E., Ravi, N., Mao, H., Rolland, C., Gustafson, L., Xiao, T., Whitehead, S., Berg, A.C., Lo, W., Doll{\'{a}}r, P., Girshick, R.B.: Segment anything. In: {IEEE/CVF} International Conference on Computer Vision, {ICCV} 2023, Paris, France, October 1-6, 2023. pp. 3992--4003. {IEEE} (2023). \doi{10.1109/ICCV51070.2023.00371}, \url{https://doi.org/10.1109/ICCV51070.2023.00371}

\bibitem{clipdino_Kobayashi_Matsumoto_Sitzmann}
Kobayashi, S., Matsumoto, E., Sitzmann, V.: Decomposing nerf for editing via feature field distillation. In: NeurIPS (2022), \url{http://papers.nips.cc/paper\_files/paper/2022/hash/93f250215e4889119807b6fac3a57aec-Abstract-Conference.html}

\bibitem{PDS}
Koo, J., Park, C., Sung, M.: Posterior distillation sampling. CoRR  \textbf{abs/2311.13831} (2023). \doi{10.48550/ARXIV.2311.13831}, \url{https://doi.org/10.48550/arXiv.2311.13831}

\bibitem{refart6_Wu_Tan_Xu_2022}
Kuang, Z., Luan, F., Bi, S., Shu, Z., Wetzstein, G., Sunkavalli, K.: Palettenerf: Palette-based appearance editing of neural radiance fields. In: {IEEE/CVF} Conference on Computer Vision and Pattern Recognition, {CVPR} 2023, Vancouver, BC, Canada, June 17-24, 2023. pp. 20691--20700. {IEEE} (2023). \doi{10.1109/CVPR52729.2023.01982}, \url{https://doi.org/10.1109/CVPR52729.2023.01982}

\bibitem{Lseg_Li_Weinberger_Belongie_Koltun_Ranftl}
Li, B., Weinberger, K.Q., Belongie, S.J., Koltun, V., Ranftl, R.: Language-driven semantic segmentation. In: The Tenth International Conference on Learning Representations, {ICLR} 2022, Virtual Event, April 25-29, 2022. OpenReview.net (2022), \url{https://openreview.net/forum?id=RriDjddCLN}

\bibitem{InstructPix2NeRF}
Li, J., Liu, S., Liu, Z., Wang, Y., Zheng, K., Xu, J., Li, J., Zhu, J.: Instructpix2nerf: Instructed 3d portrait editing from a single image. CoRR  \textbf{abs/2311.02826} (2023). \doi{10.48550/ARXIV.2311.02826}, \url{https://doi.org/10.48550/arXiv.2311.02826}

\bibitem{climatenerf}
Li, Y., Lin, Z.H., Forsyth, D., Huang, J.B., Wang, S.: Climatenerf: Extreme weather synthesis in neural radiance field. In: Proceedings of the IEEE/CVF International Conference on Computer Vision (ICCV) (2023)

\bibitem{liang2021highresollaplacian}
Liang, J., Zeng, H., Zhang, L.: High-resolution photorealistic image translation in real-time: A laplacian pyramid translation network. In: Proceedings of the IEEE/CVF Conference on Computer Vision and Pattern Recognition. pp. 9392--9400 (2021)

\bibitem{liu2023stylerf}
Liu, K., Zhan, F., Chen, Y., Zhang, J., Yu, Y., El{-}Saddik, A., Lu, S., Xing, E.P.: Stylerf: Zero-shot 3d style transfer of neural radiance fields. In: {IEEE/CVF} Conference on Computer Vision and Pattern Recognition, {CVPR} 2023, Vancouver, BC, Canada, June 17-24, 2023. pp. 8338--8348. {IEEE} (2023). \doi{10.1109/CVPR52729.2023.00806}, \url{https://doi.org/10.1109/CVPR52729.2023.00806}

\bibitem{EditNeRF_Liu_Zhang_Zhang_Zhang_Zhu_Russell_2021}
Liu, S., Zhang, X., Zhang, Z., Zhang, R., Zhu, J.Y., Russell, B.: Editing conditional radiance fields. In: 2021 IEEE/CVF International Conference on Computer Vision (ICCV) (Oct 2021). \doi{10.1109/iccv48922.2021.00572}, \url{http://dx.doi.org/10.1109/iccv48922.2021.00572}

\bibitem{SKED}
Mikaeili, A., Perel, O., Safaee, M., Cohen{-}Or, D., Mahdavi{-}Amiri, A.: {SKED:} sketch-guided text-based 3d editing. In: {IEEE/CVF} International Conference on Computer Vision, {ICCV} 2023, Paris, France, October 1-6, 2023. pp. 14561--14573. {IEEE} (2023). \doi{10.1109/ICCV51070.2023.01343}, \url{https://doi.org/10.1109/ICCV51070.2023.01343}

\bibitem{light5-Brualla_Srinivasan_Barron_2022}
Mildenhall, B., Hedman, P., Martin-Brualla, R., Srinivasan, P.P., Barron, J.T.: Nerf in the dark: High dynamic range view synthesis from noisy raw images. In: 2022 IEEE/CVF Conference on Computer Vision and Pattern Recognition (CVPR) (Jun 2022). \doi{10.1109/cvpr52688.2022.01571}, \url{http://dx.doi.org/10.1109/cvpr52688.2022.01571}

\bibitem{DBLP:conf/eccv/MildenhallSTBRN20}
Mildenhall, B., Srinivasan, P.P., Tancik, M., Barron, J.T., Ramamoorthi, R., Ng, R.: Nerf: Representing scenes as neural radiance fields for view synthesis. In: Vedaldi, A., Bischof, H., Brox, T., Frahm, J. (eds.) Computer Vision - {ECCV} 2020 - 16th European Conference, Glasgow, UK, August 23-28, 2020, Proceedings, Part {I}. Lecture Notes in Computer Science, vol. 12346, pp. 405--421. Springer (2020). \doi{10.1007/978-3-030-58452-8\_24}, \url{https://doi.org/10.1007/978-3-030-58452-8\_24}

\bibitem{brisque}
Mittal, A., Moorthy, A.K., Bovik, A.C.: No-reference image quality assessment in the spatial domain. IEEE Transactions on Image Processing p. 4695–4708 (Dec 2012). \doi{10.1109/tip.2012.2214050}, \url{http://dx.doi.org/10.1109/tip.2012.2214050}

\bibitem{Mu_Wang_Wu_Li_2022}
Mu, F., Wang, J., Wu, Y., Li, Y.: 3d photo stylization: Learning to generate stylized novel views from a single image. In: 2022 IEEE/CVF Conference on Computer Vision and Pattern Recognition (CVPR) (Jun 2022). \doi{10.1109/cvpr52688.2022.01579}, \url{http://dx.doi.org/10.1109/cvpr52688.2022.01579}

\bibitem{light4_Munkberg_Hasse}
Munkberg, J., Hasselgren, J., Shen, T., Gao, J., Chen, W., Evans, A., M\"uller, T., Fidler, S.: {Extracting Triangular 3D Models, Materials, and Lighting From Images}. In: Proceedings of the IEEE/CVF Conference on Computer Vision and Pattern Recognition (CVPR). pp. 8280--8290 (June 2022)

\bibitem{refart4_Nguyen-Phuoc_Liu_Xiao}
Nguyen{-}Phuoc, T., Liu, F., Xiao, L.: Snerf: stylized neural implicit representations for 3d scenes. {ACM} Trans. Graph.  \textbf{41}(4),  142:1--142:11 (2022). \doi{10.1145/3528223.3530107}, \url{https://doi.org/10.1145/3528223.3530107}

\bibitem{softmax_splat_Niklaus_Liu_2020}
Niklaus, S., Liu, F.: Softmax splatting for video frame interpolation. In: 2020 {IEEE/CVF} Conference on Computer Vision and Pattern Recognition, {CVPR} 2020, Seattle, WA, USA, June 13-19, 2020. pp. 5436--5445. Computer Vision Foundation / {IEEE} (2020). \doi{10.1109/CVPR42600.2020.00548}, \url{https://openaccess.thecvf.com/content\_CVPR\_2020/html/Niklaus\_Softmax\_Splatting\_for\_Video\_Frame\_Interpolation\_CVPR\_2020\_paper.html}

\bibitem{box1}
Ost, J., Mannan, F., Thuerey, N., Knodt, J., Heide, F.: Neural scene graphs for dynamic scenes. In: 2021 IEEE/CVF Conference on Computer Vision and Pattern Recognition (CVPR) (Jun 2021). \doi{10.1109/cvpr46437.2021.00288}, \url{http://dx.doi.org/10.1109/cvpr46437.2021.00288}

\bibitem{ED_NERF}
Park, J., Kwon, G., Ye, J.C.: Ed-nerf: Efficient text-guided editing of 3d scene using latent space nerf. CoRR  \textbf{abs/2310.02712} (2023). \doi{10.48550/ARXIV.2310.02712}, \url{https://doi.org/10.48550/arXiv.2310.02712}

\bibitem{clip_Radford_Kim}
Radford, A., Kim, J.W., Hallacy, C., Ramesh, A., Goh, G., Agarwal, S., Sastry, G., Askell, A., Mishkin, P., Clark, J., Krueger, G., Sutskever, I.: Learning transferable visual models from natural language supervision. In: Meila, M., Zhang, T. (eds.) Proceedings of the 38th International Conference on Machine Learning, {ICML} 2021, 18-24 July 2021, Virtual Event. Proceedings of Machine Learning Research, vol.~139, pp. 8748--8763. {PMLR} (2021), \url{http://proceedings.mlr.press/v139/radford21a.html}

\bibitem{VGG}
Simonyan, K., Zisserman, A.: Very deep convolutional networks for large-scale image recognition. arXiv preprint arXiv:1409.1556  (2014)

\bibitem{DBLP:journals/corr/SimonyanZ14a}
Simonyan, K., Zisserman, A.: Very deep convolutional networks for large-scale image recognition. In: Bengio, Y., LeCun, Y. (eds.) 3rd International Conference on Learning Representations, {ICLR} 2015, San Diego, CA, USA, May 7-9, 2015, Conference Track Proceedings (2015), \url{http://arxiv.org/abs/1409.1556}

\bibitem{Blending-NeRF}
Song, H., Choi, S., Do, H., Lee, C., Kim, T.: Blending-nerf: Text-driven localized editing in neural radiance fields. In: {IEEE/CVF} International Conference on Computer Vision, {ICCV} 2023, Paris, France, October 1-6, 2023. pp. 14337--14347. {IEEE} (2023). \doi{10.1109/ICCV51070.2023.01323}, \url{https://doi.org/10.1109/ICCV51070.2023.01323}

\bibitem{light3_Mildenhall_Barron_2021}
Srinivasan, P.P., Deng, B., Zhang, X., Tancik, M., Mildenhall, B., Barron, J.T.: Nerv: Neural reflectance and visibility fields for relighting and view synthesis. In: 2021 IEEE/CVF Conference on Computer Vision and Pattern Recognition (CVPR) (Jun 2021). \doi{10.1109/cvpr46437.2021.00741}, \url{http://dx.doi.org/10.1109/cvpr46437.2021.00741}

\bibitem{nerfstudio}
Tancik, M., Weber, E., Ng, E., Li, R., Yi, B., Kerr, J., Wang, T., Kristoffersen, A., Austin, J., Salahi, K., Ahuja, A., McAllister, D., Kanazawa, A.: Nerfstudio: A modular framework for neural radiance field development. In: ACM SIGGRAPH 2023 Conference Proceedings. SIGGRAPH '23 (2023)

\bibitem{flow_Teed_Deng_2020}
Teed, Z., Deng, J.: {RAFT:} recurrent all-pairs field transforms for optical flow (extended abstract). In: Zhou, Z. (ed.) Proceedings of the Thirtieth International Joint Conference on Artificial Intelligence, {IJCAI} 2021, Virtual Event / Montreal, Canada, 19-27 August 2021. pp. 4839--4843. ijcai.org (2021). \doi{10.24963/IJCAI.2021/662}, \url{https://doi.org/10.24963/ijcai.2021/662}

\bibitem{clipdino_Tschernezki_Laina_Larlus_Vedaldi}
Tschernezki, V., Laina, I., Larlus, D., Vedaldi, A.: Neural feature fusion fields: 3d distillation of self-supervised 2d image representations. In: International Conference on 3D Vision, 3DV 2022, Prague, Czech Republic, September 12-16, 2022. pp. 443--453. {IEEE} (2022). \doi{10.1109/3DV57658.2022.00056}, \url{https://doi.org/10.1109/3DV57658.2022.00056}

\bibitem{instnorm}
Ulyanov, D., Vedaldi, A., Lempitsky, V.S.: Instance normalization: The missing ingredient for fast stylization. CoRR  \textbf{abs/1607.08022} (2016), \url{http://arxiv.org/abs/1607.08022}

\bibitem{light1_Verbin_Hedman_Mildenhall_Zicklen_2022}
Verbin, D., Hedman, P., Mildenhall, B., Zickler, T., Barron, J.T., Srinivasan, P.P.: Ref-nerf: Structured view-dependent appearance for neural radiance fields. In: 2022 IEEE/CVF Conference on Computer Vision and Pattern Recognition (CVPR) (Jun 2022). \doi{10.1109/cvpr52688.2022.00541}, \url{http://dx.doi.org/10.1109/cvpr52688.2022.00541}

\bibitem{ClipNeRF_Wang_Chai_He_Chen_Liao_2022}
Wang, C., Chai, M., He, M., Chen, D., Liao, J.: Clip-nerf: Text-and-image driven manipulation of neural radiance fields. In: 2022 IEEE/CVF Conference on Computer Vision and Pattern Recognition (CVPR) (Jun 2022). \doi{10.1109/cvpr52688.2022.00381}, \url{http://dx.doi.org/10.1109/cvpr52688.2022.00381}

\bibitem{NeRFArt_Wang_Jiang_Chai_He_Chen_Liao_2022}
Wang, C., Jiang, R., Chai, M., He, M., Chen, D., Liao, J.: Nerf-art: Text-driven neural radiance fields stylization. IEEE Transactions on Visualization and Computer Graphics pp. 1--15 (2023). \doi{10.1109/TVCG.2023.3283400}

\bibitem{PhysGaussian}
Xie, T., Zong, Z., Qiu, Y., Li, X., Feng, Y., Yang, Y., Jiang, C.: Physgaussian: Physics-integrated 3d gaussians for generative dynamics. CoRR  \textbf{abs/2311.12198} (2023). \doi{10.48550/ARXIV.2311.12198}, \url{https://doi.org/10.48550/arXiv.2311.12198}

\bibitem{box2}
Yu, H., Guibas, L.J., Wu, J.: Unsupervised discovery of object radiance fields. In: The Tenth International Conference on Learning Representations, {ICLR} 2022, Virtual Event, April 25-29, 2022. OpenReview.net (2022), \url{https://openreview.net/forum?id=rwE8SshAlxw}

\bibitem{box_deform}
Yuan, Y., Sun, Y., Lai, Y., Ma, Y., Jia, R., Gao, L.: Nerf-editing: Geometry editing of neural radiance fields. In: {IEEE/CVF} Conference on Computer Vision and Pattern Recognition, {CVPR} 2022, New Orleans, LA, USA, June 18-24, 2022. pp. 18332--18343. {IEEE} (2022). \doi{10.1109/CVPR52688.2022.01781}, \url{https://doi.org/10.1109/CVPR52688.2022.01781}

\bibitem{refart5_Zhang_Kolkin_Bi}
Zhang, K., Kolkin, N.I., Bi, S., Luan, F., Xu, Z., Shechtman, E., Snavely, N.: {ARF:} artistic radiance fields. In: Avidan, S., Brostow, G.J., Ciss{\'{e}}, M., Farinella, G.M., Hassner, T. (eds.) Computer Vision - {ECCV} 2022 - 17th European Conference, Tel Aviv, Israel, October 23-27, 2022, Proceedings, Part {XXXI}. Lecture Notes in Computer Science, vol. 13691, pp. 717--733. Springer (2022). \doi{10.1007/978-3-031-19821-2\_41}, \url{https://doi.org/10.1007/978-3-031-19821-2\_41}

\bibitem{lpips_Zhang_Isola_Efros_Shechtman_Wang_2018}
Zhang, R., Isola, P., Efros, A.A., Shechtman, E., Wang, O.: The unreasonable effectiveness of deep features as a perceptual metric. In: 2018 IEEE/CVF Conference on Computer Vision and Pattern Recognition (Jun 2018). \doi{10.1109/cvpr.2018.00068}, \url{http://dx.doi.org/10.1109/cvpr.2018.00068}

\bibitem{DreamEditor}
Zhuang, J., Wang, C., Liu, L., Lin, L., Li, G.: Dreameditor: Text-driven 3d scene editing with neural fields. CoRR  \textbf{abs/2306.13455} (2023). \doi{10.48550/ARXIV.2306.13455}, \url{https://doi.org/10.48550/arXiv.2306.13455}

\end{thebibliography}
\end{document}